\documentclass[11pt]{article}

\usepackage[dvipsnames,table]{xcolor}
\usepackage[final]{acl}

\usepackage{times}
\usepackage{latexsym}
\usepackage{array}

\usepackage[T1]{fontenc}
\usepackage{pgfplots}
\pgfplotsset{compat=1.18}
\usepackage[utf8]{inputenc}

\usepackage{microtype}
\usepackage{array}
\usepackage{inconsolata}

\usepackage{graphicx}
\usepackage{fix-cm}
\usepackage{graphicx}
\usepackage{amsfonts}
\usepackage{amsmath}
\usepackage{amssymb}
\usepackage{cleveref}
\usepackage{nicefrac}
\usepackage{xspace}
\usepackage{enumitem}
\usepackage{wrapfig}
\usepackage[most]{tcolorbox}
\usepackage{listings}
\usepackage{multirow}
\usepackage{multicol}
\usepackage{siunitx}
\usepackage{subcaption}
\usepackage{nicematrix}
\usepackage{makecell}
\usepackage{booktabs} 
\title{DeepResearch RewardBench: Evaluating Reward Models for Deep Research Report}

\author{%
Shudong Liu$^{1,}$\thanks{
Equal contribution.
$^{\dagger}$Corresponding authors.},
Yuting Zhong$^{1,*}$,
Guanhua Chen$^{1}$,
Xinyi Yang$^{1}$,
\\
\textbf{Junnan Liu$^{2}$,
Lidia S. Chao$^{1}$,
Jiaheng Liu$^{3}$,
Derek F. Wong$^{1,\dagger}$
}
\\
\textsuperscript{$^{1}$}NLP$^{2}$CT Lab, University of Macau \quad \textsuperscript{$^{2}$}Monash University \quad \textsuperscript{$^{3}$}Nanjing University\\
\texttt{nlp2ct.\{shudong,yuting,guanhua,xinyi\}@gmail.com; \{derekfw,lidiasc\}@um.edu.mo}
}

\begin{document}
\maketitle
\begin{abstract}
Empowered by Large Language Models (LLMs), deep research systems integrate online information into research reports through multi-step web searching and information synthesis. Evaluating these reports presents a significant challenge due to their long context, high information density, and domain specificity, necessitating the use of AI evaluation as a substitute for prohibitively expensive human assessment. However, few studies have examined the effectiveness of existing reward models or LLMs in understanding and evaluating such research reports. To bridge this gap, we introduce \textbf{DeepResearch RewardBench}, the first benchmark dedicated to evaluating reward modeling for research reports. We collect diverse report candidates from multiple agents or tool-augmented LLMs. These candidates are scored using query-specific golden rubrics, and we ensure data quality and difficulty through a combination of multi-round filtering and human verification. Our benchmark comprises preference datasets formatted as pairs, triplets, and quadruplets to test ranking capabilities across varying levels of difficulty. Extensive experiments demonstrate that while current reward models handle general coherence effectively, their performance significantly degrades when evaluating research reports. Generative LLMs acting as judges demonstrate comprehensive superiority in this domain, with comparative evaluation proving more effective than direct scoring methods. We anticipate that this benchmark will serve as a vital testbed for the evaluation and advancement of research agents\footnote{Code and data can be found at: \url{https://github.com/NLP2CT/DRRB}}.

\end{abstract}
\section{Introduction}
Reward Models (RMs) are essential for aligning Large Language Models (LLMs) with human values, serving as proxies for human preference in Reinforcement Learning from Human Feedback (RLHF)~\cite{zhong2025comprehensive,liu2025compassverifier}. Recently, the emergence of autonomous research agents has shifted the focus from daily dialogue to Deep Research, where models must autonomously search, read, and synthesize information into comprehensive reports \cite{du2025deepresearch,xu2025researcherbench}. This paradigm shift raises a critical question: can existing RMs reliably evaluate such complex, long-form research outputs?

Although benchmarks like RewardBench v1/v2~\cite{lambert2025rewardbench,malik2025rewardbench} have been developed for preference modeling evaluation, they remain insufficient for the Deep Research domain. While existing benchmarks primarily focus on chat-style coherence and safety alignment, Deep Research reports present more demanding challenges, such as information redundancy inherent in long-form content, specialized scientific knowledge, and the integration of extensive citations. Existing evaluation methods primarily rely on designing extensive, specific rubrics and retrieving external information for pointwise scoring. However, for complex reports that lack absolute objectivity, pairwise comparison serves as a more suitable evaluation paradigm.

\begin{figure*}[t]
    \centering 
    \includegraphics[width=.96\linewidth]{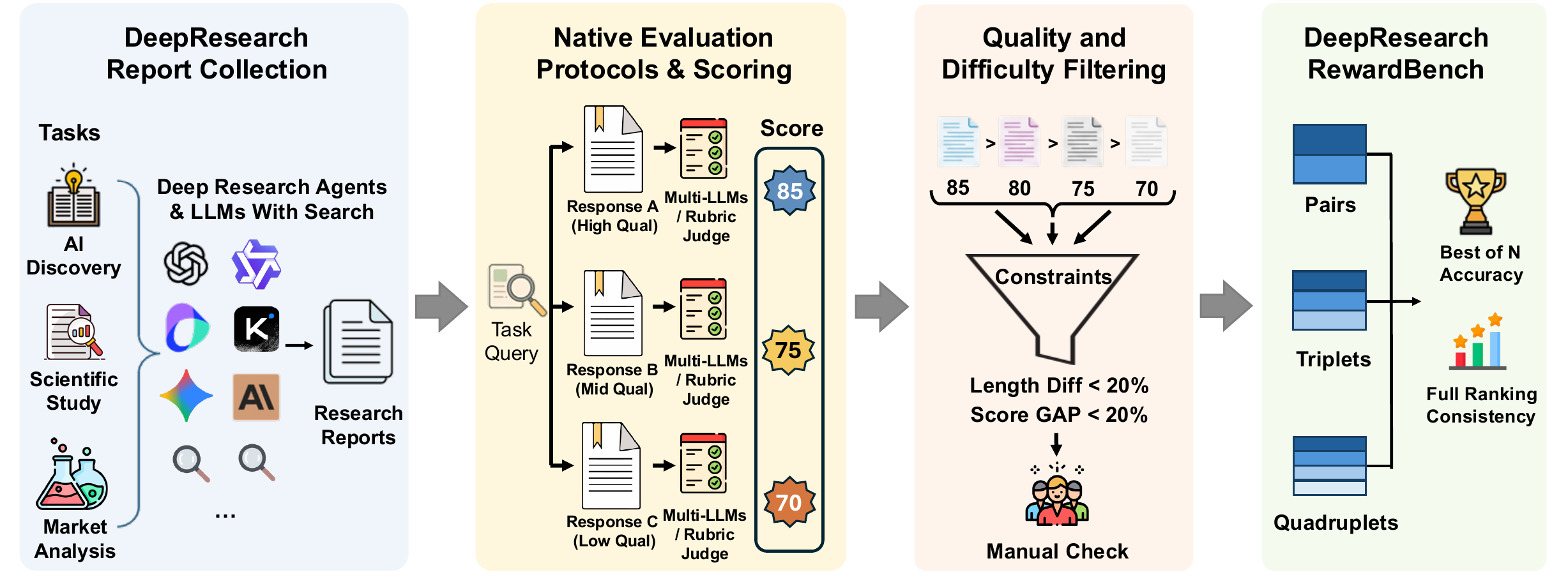}
    \vspace{-0.5em} 
    \caption{Overview of the DeepResearch RewardBench's data collection, scoring, screening, and evaluation process.}
    \label{fig-dataset}
\end{figure*}

To address these limitations, we introduce \textbf{DeepResearch RewardBench}, a benchmark designed to evaluate RMs' capabilities in judging deep research reports. We build our dataset upon mainstream Deep Research datasets and design a pipeline that achieves accurate annotation. 
Specifically, we first aggregate responses generated by various agents and LLMs, then score them following the evaluation protocols provided by the original benchmarks. These protocols, developed by the benchmark authors, incorporate task-specific evaluation rubrics. After filtering for quality and difficulty, the final dataset is cross-checked by PhD-level human annotators.

We evaluate representative RMs, including sequence classifiers, generative LLMs-as-a-Judge, and DPO-based RMs, on DeepResearch RewardBench.
Our results demonstrate that without task-specific tuning, generative LLMs exhibit superior performance and robustness, whereas traditional reward models (RMs) face a clear ceiling. Furthermore, we conduct a deeper analysis and a human-centric error taxonomy on generative LLMs. In the absence of a golden rubric, the LLM-as-a-Judge approach under a comparative framework outperforms pointwise scoring. Nonetheless, common failure modes persist, such as selecting inappropriate assessment dimensions during evaluation, and the judge models being constrained by a limited knowledge base, which hinders their comprehensive evaluation of long-form reports or those requiring up-to-date information.

Our contributions are as follows: 1) we introduce DeepResearch RewardBench, the first benchmark dedicated to evaluating reward modeling on deep research reports; 2) we propose a data construction pipeline that combines golden-rubric-based filtering with human verification to ensure highly reliable preference labels; and 3) we conduct a comprehensive evaluation across varying tuple sizes and analyze an error taxonomy, revealing critical limitations in current RMs' ability to meet rigorous research standards.

\section{Related Work}
\paragraph{Reward Model Benchmark} RLHF has become a dominant alignment framework that trains reward models on human preference data to optimize LLMs~\cite{dai2023safe,chaudhari2025rlhf}. RMs can be broadly categorized into three types: sequence classifiers that append a regression head to predict scores, Direct Preference Optimization~\cite{rafailov2023direct} (DPO) methods that reparameterize preference-based rewards using only policy models, and generative approaches that leverage LLMs as judges to evaluate completions. Early reward model benchmarks such as RewardBench~\cite{lambert2025rewardbench} and RM-bench~\citep{liu2025rm} formulated evaluation as classification tasks across dialogue, reasoning, and safety domains. Subsequent work extended this to additional dimensions, including factuality and instruction following~\cite{malik2025rewardbench}, as well as multilingual settings~\cite{gureja2025m} and multimodel settings~\cite{li2025vlrewardbench}. While existing RewardBenchs typically evaluate general and subjective chat content, such as creative writing and safety alignment, by comparing responses from different models, they fall short for deep research reports. These reports feature long-form contexts, high information density, and highly complex evaluation dimensions, presenting a critical gap in current retrieval and evaluation benchmarks.

\begin{figure}[t!]
    \centering \includegraphics[width=.96\linewidth]{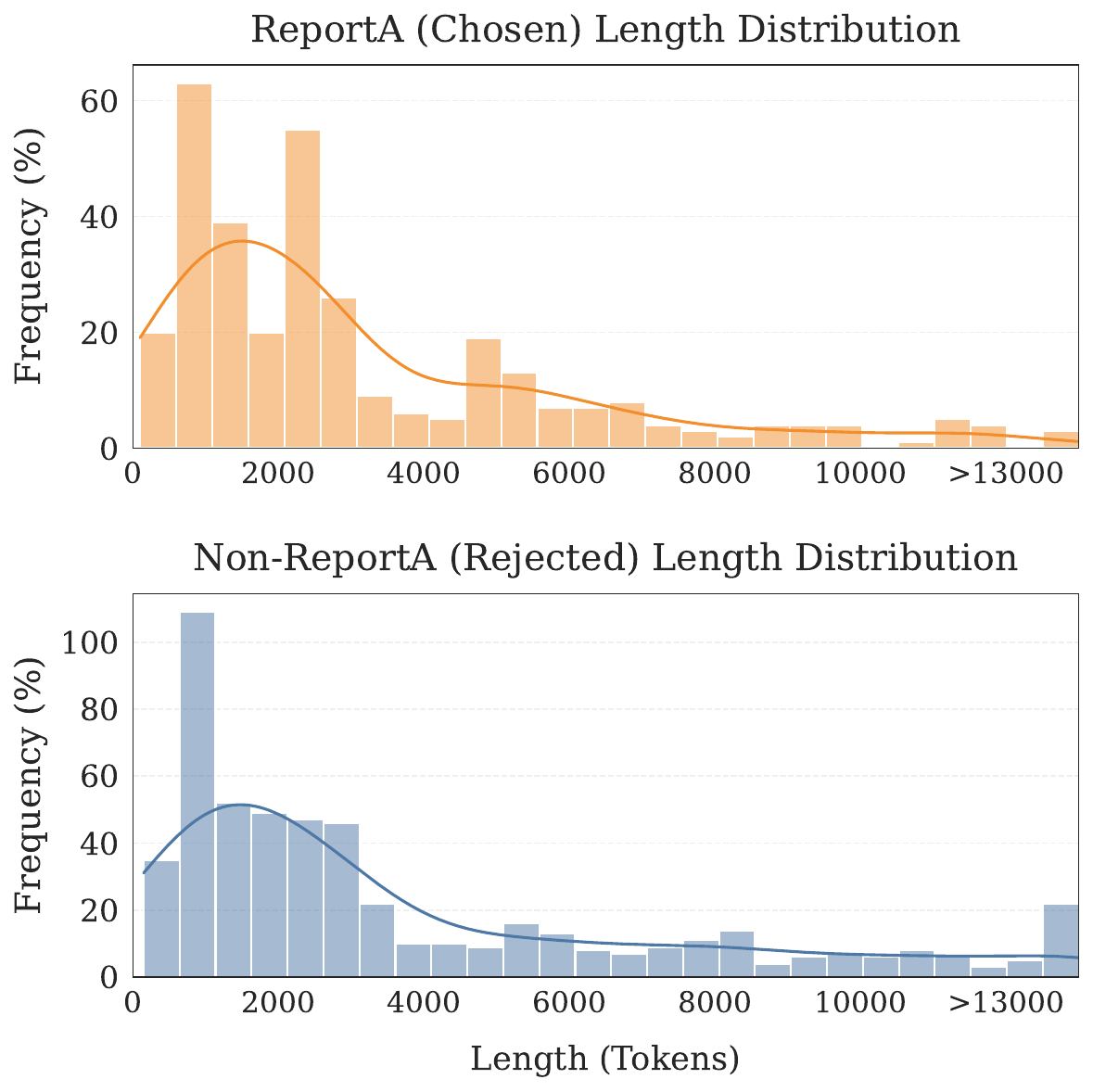}
    \vspace{-0.7em} 
    \caption{Report length distribution in DeepResearch RewardBench.}
    \label{fig-length-distribution}
\end{figure}

\paragraph{Deep Research Evaluation} Meanwhile, several benchmarks have emerged for evaluating Deep Research systems. DeepResearchBench~\cite{du2506deepresearch} uses FACT-based and RACE-based metrics on PhD-curated tasks, while DRBench~\cite{abaskohi2025drbench} assesses insight, recall, and factuality in industrial settings. Other work explores component-aware evaluation~\cite{chen2508browsecomp}, claim-based precision and recall~\cite{java2025characterizing}, and literature-centric assessment~\cite{patel2025deepscholar,sharma2025researchrubrics}. Recent efforts also investigate automated judging paradigms~\cite{wan2025deepresearch,xu2025researcherbenchevaluatingdeepai} and personalized evaluation~\cite{liang2025towards,gou2025mind2web}. Despite this progress, evaluating research reports remains challenging due to the absence of gold standards and the multi-dimensional nature of quality assessment. However, training LLMs, such as through (Rejected-sampling) Supervised Fine-Tuning or RLHF, demands large-scale evaluation that renders exclusive reliance on human experts prohibitively expensive. Therefore, most benchmarks rely on LLM-as-a-judge approaches, yet it remains unclear how well current reward models can evaluate long-form research outputs. Our work addresses this gap by constructing a benchmark specifically designed to assess reward model performance on Deep Research reports.

\section{DeepResearch RewardBench}
Figure \ref{fig-dataset} illustrates the comprehensive dataset construction pipeline and evaluation methodology of DeepResearch RewardBench, organized into the following primary stages.
\subsection{Benchmark Construction Pipeline}
\paragraph{Research Report Collection.} 
We curated a diverse set of research tasks from two prominent long-form-report Deep Research datasets, DeepResearch Bench and Researcher Bench~\cite{du2025deepresearch,xu2025researcherbench}, spanning various domains such as AI Research, Finance, and Technology. To ensure a broad distribution of response styles and capabilities, we collected research reports generated by SOTA Deep Research Agents (e.g., Deep Research Agents from Gemini, OpenAI, Claude, Doubao, Kimi) as well as general LLMs equipped with web search tools (e.g., GPT-4o, Claude 3.5 Sonnet)~\cite{2410.21276,comanici2025gemini,anthropic2025claudehaiku}. We strictly adhere to the specific experimental settings of each dataset to ensure the correctness and reproducibility of the report.

\begin{figure}[t!]
    \centering \includegraphics[width=.96\linewidth]{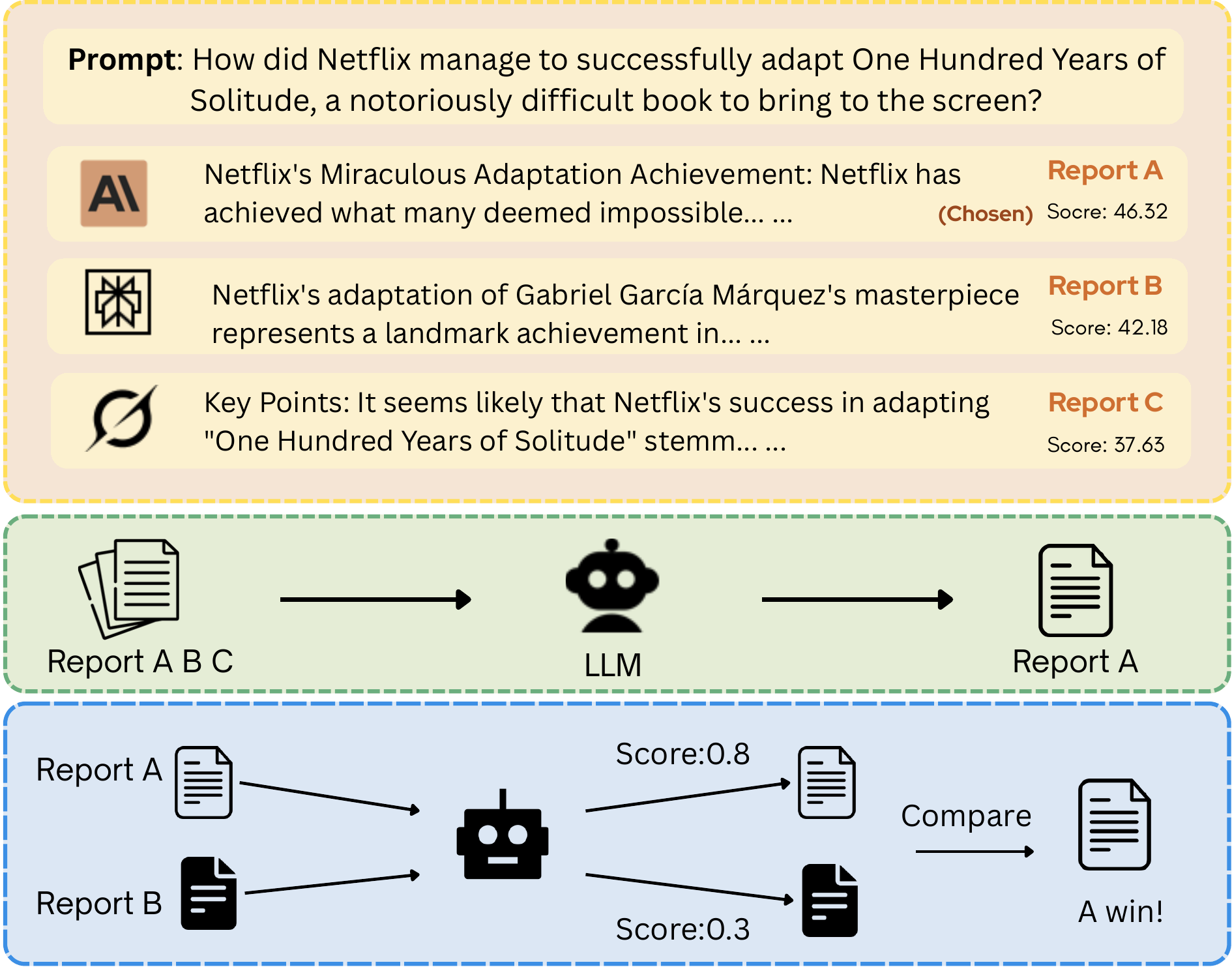}
    \vspace{-0.5em}
    \caption{Sample and evaluation tasks in DeepResearch RewardBench.}
    \label{fig-dataset-example}
\end{figure}

\paragraph{Adopting Native Evaluation Protocols for Evaluation.}
The developers of these deep research benchmarks have established sophisticated, customized evaluation frameworks tailored specifically to their respective datasets. These frameworks have undergone rigorous human verification, offering significantly higher precision than general-purpose tools and achieving accuracy comparable to human annotation. Consequently, we fully adopted their original evaluation protocols, including specific scoring rubrics and designated judge models (Gemini-2.5-Pro and o3-mini), to score the reports for each dataset. This ensures that our quality assessment remains grounded in the intrinsic requirements of deep research tasks.
{\color{black}Specifically, for DeepResearch Bench~\cite{du2025deepresearch}, we adopt the \textbf{RACE} (Reference-based Adaptive Criteria-driven Evaluation) framework, which evaluates reports across four orthogonal dimensions: \emph{Comprehensiveness}, \emph{Insight/Depth}, \emph{Instruction-Following}, and \emph{Readability}. RACE dynamically generates task-specific dimension weights and fine-grained criteria, then employs reference-based scoring with Gemini-2.5-Pro as the Judge LLM. For ResearcherBench~\cite{xu2025researcherbench}, we adopt its \textbf{rubric assessment} framework, where expert-designed evaluation rubric items with assigned importance weights are used to measure the coverage of key insights in each report, with o3-mini as the Judge LLM. Detailed descriptions and examples of both evaluation protocols are provided in Appendix~\ref{appendix_rubrics}.}

\paragraph{Data Pairing and Filtering.}
To balance the difficulty and quality of the benchmark, we performed systematic pairing and filtering based on report length and scoring margins, resulting in three distinct subsets: pairs (2-tuples), triplets (3-tuples), and quadruplets (4-tuples).
{\color{blue}}Each tuple consists of reports generated for the \emph{same} research query by different agents or LLMs. The construction procedure is as follows: for each query, we first collect all available candidate reports and sort them by their evaluation scores. We then enumerate all possible $k$-combinations ($k \in \{2, 3, 4\}$) from these sorted candidates. Each combination is filtered by 
1) \textbf{\emph{Score Consistency.}} Reports in each tuple are arranged by score gradient from high to low, where the top-ranked report represents the \emph{Chosen} sample and the others function as \emph{Rejected}. For DeepResearch Bench~\cite{du2025deepresearch}, which comprises four sub-scores and a weighted average, we selected instances where at least three sub-scores align with the trend of the average score. For ResearcherBench, which originally provides only a single metric, we employed K2-Thinking~\cite{kimi-k2-techreport-2025} for supplementary scoring and filtered for consistency with o3-mini trends.
2) \textbf{\emph{Length Proximity.}} Since different agents and LLMs produce reports with varying verbosity, we first applied regex-based sanitization to remove all URLs from the reports, preventing an abundance of links from skewing length measurements. We then filtered out candidates with excessive length discrepancies, retaining only those groups where the length variation (after URL removal) was within a 10\% threshold. 3) \textbf{\emph{Score Margin.}} To control for task difficulty and mitigate easy comparisons, we excluded pairs with a score margin greater than 10/30 points{\color{black}, defined as the absolute difference between the highest and lowest report scores within a tuple on a 100-point scale}. The RACE scores for DeepResearch Bench generally fall within a relatively narrow range because they are comparative scores averaged across multiple dimensions, whereas ResearcherBench usually computes a weighted score from only a small number of rubric items, so a difference of one or two items can lead to a much larger score gap. We therefore use a 10\% threshold for DeepResearch Bench and a 30\% threshold for ResearcherBench. This rigorous selection process ensures that DeepResearch RewardBench minimizes length and URL biases while maintaining a high level of discriminative challenge.

\paragraph{Preference Validation and Quality Assurance.}
We employs a two-tiered safeguard mechanism to ensure this quality: 1) \emph{Validating Evaluation Standards}. We utilize the specific rubrics and recommended Judge Models associated with each dataset. These rubrics have undergone expert human verification to ensure accuracy, while the selected Judge Models have demonstrated superior performance across various LLMs. This rigorous selection process enables AI evaluations to achieve high agreement with expert annotations, even surpassing the inter-annotator agreement observed among human experts. 2) \emph{Human-in-the-Loop Verification}: Following the formation of tuples, we conducted manual verification to filter out instances where human assessment diverged completely from the AI evaluation. All human evaluators hold PhD degrees or are PhD students and have performed comparative analyses strictly grounded in the established rubrics and reports. Details are in Section \ref{Annotation Details}.

\begin{table}[t!]
\centering
\resizebox{0.4\textwidth}{!}{
\begin{tabular}{lccc}
\toprule
\textbf{Subset} & \textbf{\# Instances (Tasks)} & \textbf{\# Reports} \\ \midrule
Pairs       & 160 & 320 \\
Triplets    & 138 & 414 \\
Quadruplets & 33  & 132 \\ \midrule
Total & 331 & 866 \\ \bottomrule
\end{tabular}}
\vspace{-0.7em}
\caption{Data statistics of DeepResearch RewardBench. Each instance represents a unique research task with multiple candidate reports for ranking.}
\label{tab:data_stats}
\end{table}

\subsection{Data Statistics}
Following careful data screening and filtering, DeepResearch RewardBench comprises 331 instances, encompassing a total of 866 reports for comparison. Detailed data statistics are presented in Table \ref{tab:data_stats}. Deep Research is still in its early stages. Since every research task (query) must provide practical value and include corresponding evaluation techniques, the overall volume remains relatively small. Furthermore, due to our stringent constraints on length consistency and task difficulty, coupled with multiple rounds of filtering, the number of quadruplet instances is relatively smaller.

\paragraph{Length Distribution} Figure \ref{fig-length-distribution} illustrates the length distribution of selected versus rejected reports, tokenized using the Qwen2.5-7B-Instruct tokenizer. As observed, the research reports generally maintain a long-text profile ranging between 1,000 and 5,000 tokens, with a few ultra-long instances exceeding 10k tokens. Notably, the rejected reports are occasionally even longer than the selected ones; this is due to our deliberate control over length proximity during the data construction process. We aim to counteract the common ``length bias'', where traditional RMs tend to favor longer responses.

\begin{figure}[t!]
    \centering
    \includegraphics[width=0.27\textwidth]{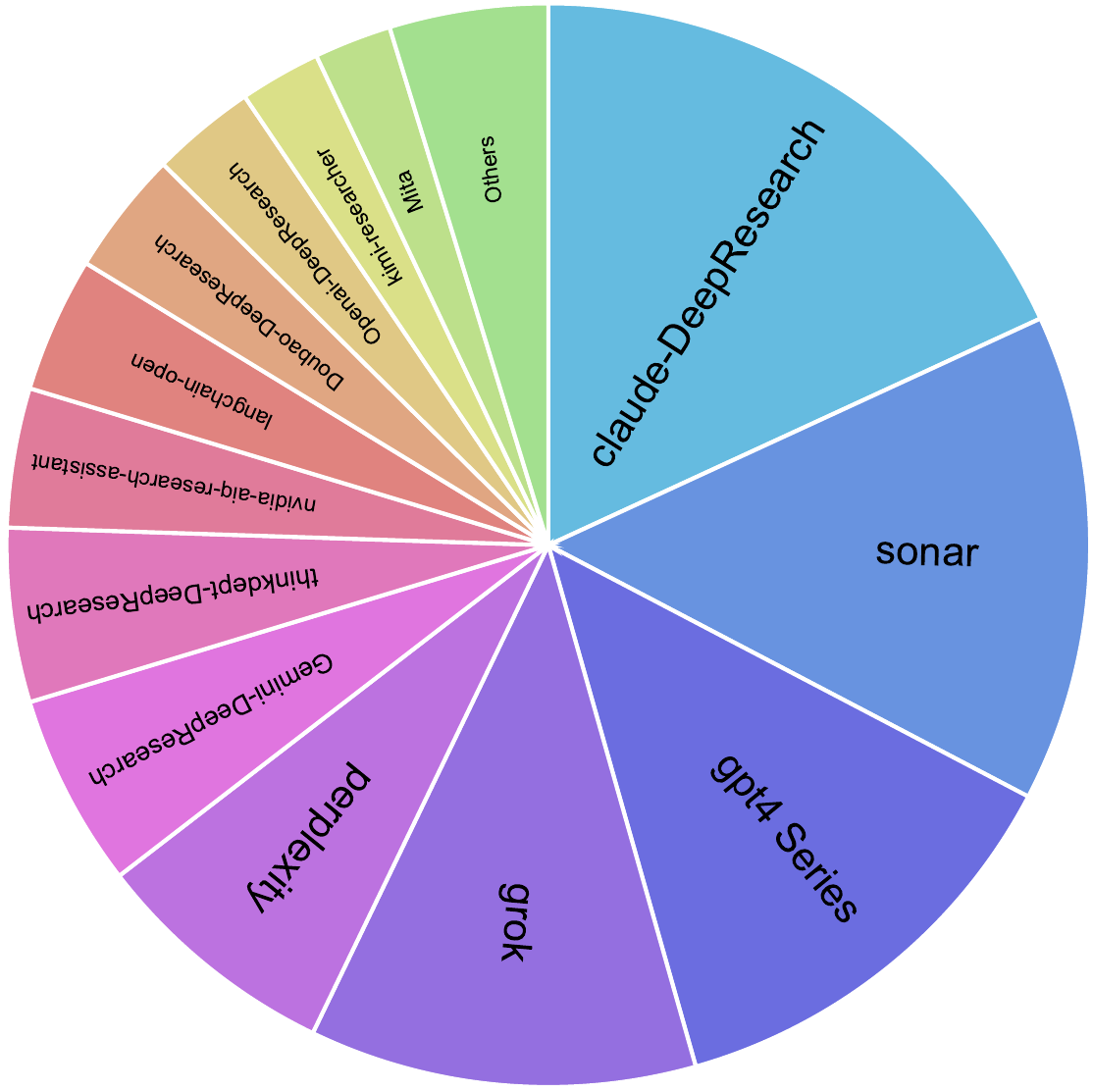}
    \vspace{-0.7em}
    \caption{Distribution of model usage in research reports. 
    Approximate models from the same family were merged together. 
    }
    \label{fig:model_distribution}
\end{figure}

\paragraph{Model Distribution} A detailed breakdown of the models utilized to generate these research reports is illustrated in figure~\ref{fig:model_distribution}. This includes 19 different families of agents, where models from the same series within a family are grouped.

\definecolor{bgGen}{RGB}{235, 245, 255} 
\definecolor{bgSeq}{RGB}{240, 255, 240} 
\definecolor{bgDPO}{RGB}{255, 245, 235} 

\begin{table*}[t!]
    \centering

    \resizebox{.92\textwidth}{!}{%
    \begin{tabular}{lccccccc}
        \toprule
        \multirow{2}{*}{\textbf{Models}} & \multirow{2}{*}{\textbf{Pair}} & \multicolumn{2}{c}{\textbf{Triplet}} & \multicolumn{2}{c}{\textbf{Quadruplet}} & \multirow{2}{*}{\textbf{Average}} & \multirow{2}{*}{\textbf{Types}} \\
        \cmidrule(lr){3-4} \cmidrule(lr){5-6} 
        
         & & \textbf{BoN} & \textbf{Rank} & \textbf{BoN} & \textbf{Rank} & & \\
        \midrule
        
        \rowcolor{bgGen} google/gemini-2.5-flash	& 81.87	& 91.30	& 70.28	& 81.25	& 31.25	& 78.87 & Generative \\
        \rowcolor{bgGen} Qwen/Qwen3-235B-A22B-Instruct-2507 & 81.25 & 84.78 & 66.66 & 78.79 & 60.61 & 77.79 & Generative \\
        \rowcolor{bgGen} Qwen/Qwen3-235B-A22B-Thinking-2507 & 80.62 & 83.33 & 70.29 & 87.88 & 48.48 & 77.79 & Generative \\
        \rowcolor{bgGen} moonshotai/Kimi-K2-Instruct-0905	& 87.50	& 79.71	& 61.59	& 66.66	& 39.39	& 77.04 & Generative \\
        \rowcolor{bgGen} openai/gpt-4.1	& 91.87	& 60.14	& 47.10	& 57.57	& 33.33	& 71.29 & Generative \\
        \rowcolor{bgGen} deepseek-ai/DeepSeek-R1-0528 & 75.00 & 72.46 & 54.35 & 57.58 & 21.21 & 66.62 & Generative \\
        \rowcolor{bgGen} deepseek-ai/DeepSeek-V3-0324 & 80.00 & 62.31 & 42.75 & 54.54 & 21.21 & 64.35 & Generative \\
        
        \rowcolor{bgSeq} allenai/Llama-3.1-70B-Instruct-RM-RB2 & 66.25 & 69.57 & 53.62 & 72.73 & 48.48 & 63.75 & Seq Classifiers \\
        \rowcolor{bgSeq} Skywork/Skywork-Reward-V2-Llama-3.1-8B & 68.12 & 66.67 & 47.83 & 66.67 & 30.30 & 61.63 & Seq Classifiers \\
        \rowcolor{bgSeq} Skywork/Skywork-Reward-V2-Llama-3.2-3B & 64.38 & 65.22 & 50.00 & 66.67 & 27.27 & 59.82 & Seq Classifiers \\
        \rowcolor{bgSeq} RLHFlow/ArmoRM-Llama3-8B-v0.1 & 58.75 & 63.04 & 44.93 & 57.58 & 30.30 & 55.29 & Seq Classifiers \\
        \rowcolor{bgSeq} hendrydong/Mistral-RM-for-RAFT-GSHF-v0 & 65.62 & 52.17 & 36.23 & 57.58 & 30.30 & 54.53 & Seq Classifiers \\
        \rowcolor{bgSeq} weqweasdas/RM-Mistral-7B & 65.00 & 50.72 & 36.96 & 54.54 & 36.36 & 54.23 & Seq Classifiers \\
        \rowcolor{bgSeq} Ray2333/reward-model-Mistral-7B-Unified & 60.51 & 52.17 & 36.96 & 51.52 & 24.24 & 51.61 & Seq Classifiers \\
        \rowcolor{bgSeq} sfairXC/FsfairX-LLaMA3-RM-v0.1 & 63.75 & 49.28 & 29.71 & 36.36 & 12.12 & 49.70 & Seq Classifiers \\
        
        \rowcolor{bgDPO} 0-hero/Matter-0.1-7B-boost-DPO-preview & 44.38 & 42.03 & 19.57 & 39.39 & 9.09 & 36.71 & DPO \\
        
        \rowcolor{bgGen} prometheus-eval/prometheus-7b-v2.0 & 50.32 & 31.16 & 20.29 & 18.18 & 9.09 & 36.41 & Generative \\
        
        \rowcolor{bgDPO} NousResearch/Nous-Hermes-2-Mistral-7B-DPO & 40.88 & 28.26 & 12.32 & 48.48 & 9.09 & 31.09 & DPO \\
        
        \rowcolor{bgSeq} RLHFlow/pair-preference-model-LLaMA3-8B & 38.75 & 18.84 & 7.25 & 30.30 & 0.00 & 25.68 & Seq Classifiers \\
        
        \bottomrule
    \end{tabular}%
    }
    \vspace{-0.6em}
    \caption{Model Performance Comparison on DeepResearch RewardBench. Rows are colored by type: \colorbox{bgGen}{Generative}, \colorbox{bgSeq}{Seq Classifiers}, and \colorbox{bgDPO}{DPO}. For pairs, BoN and Rank metrics are equivalent. {\color{black}The final average is the weighted mean of the five column scores (Pair, Triplet-BoN, Triplet-Rank, Quadruplet-BoN, Quadruplet-Rank), weighted by the number of instances in each subset.
    }}
    \label{tab:model_performance}
\end{table*}

\section{Experiments}
\subsection{Evaluation Setup}
\paragraph{Metrics}
To rigorously evaluate the performance of reward models across our three subsets, we employ two primary metrics: Best-of-N (BoN) Selection and Full Ranking Consistency.
\begin{itemize}[leftmargin=1em]
 \item \textbf{Best-of-N Selection Accuracy}: For all subsets (pairs, triplets, and quadruplets), we assess whether the RMs can successfully identify the highest-quality candidate. A model’s prediction is considered correct only if it assigns the highest scalar score to the chosen report relative to all other rejected candidates in the tuple. 
 \item \textbf{Full Ranking Consistency}: To further investigate the models' discriminative granularity, we introduce a more challenging ranking task for the triplets and quadruplets. In this setup, the RMs must assign scores or provide a sorting order that reflect the exact preference order established by our gold-standard labels. A prediction is marked as correct only if the model's score sequence is perfectly monotonic with the ground-truth ranking. This task provides a more stringent test of the model's ability to capture subtle quality differences across multiple research reports.
\end{itemize}

\paragraph{Implementation Details}
For all models, we employ temperature=0.01 for evaluation/verification, with both $max\_gen\_len$ and $max\_model\_len$ set to their maximum values. 
Since most open-source LLMs cannot accommodate the input length required for comparing more than two reports simultaneously, we employ more powerful proprietary models for our generative reward modeling experiments. For the small fraction of reports (less than 1\%) that still exceed the maximum context window of sequence-classifier reward models, we pre-tokenize the text and truncate the final portion of the report to satisfy length constraints. This approach is justified as the end of such reports typically consists of an excessive amount of URLs.

\paragraph{Evaluated Models}
 
We conduct comprehensive evaluations on DeepResearch RewardBench, the first benchmark specifically designed for evaluating reward models on deep research reports. Our evaluation encompasses 19 models spanning diverse architectures and scales from 7B to 235B parameters, categorized into three groups: 
\begin{itemize}[leftmargin=1em]
 \item \textbf{Generative LLMs} including Qwen3-235B-A22B-Instruct and Qwen3-235B-A22B-Thinking \cite{qwen3-techreport}, DeepSeek-R1 \cite{deepseek-r1-2025}, DeepSeek-V3 \cite{deepseek-v3-techreport-2024}, Kimi-K2-Instruct \cite{kimi-k2-techreport-2025}, GPT-4.1 \cite{openai-gpt41-docs}, Gemini-2.5-Flash \cite{google-gemini25-flash-docs}, and Prometheus-7b-v2.0 \cite{prometheus2-2024}
 \item \textbf{Sequence classifier-based reward models} such as Skywork-Reward-V2-Llama-3.1-8B and Skywork-Reward-V2-Llama-3.2-3B \cite{skywork-reward-v2-2025}, ArmoRM-Llama3-8B-v0.1 \cite{armorm-emnlp-2024}, FsfairX-LLaMA3-RM-v0.1 \cite{raft-2023,iterative-pref-learning-2024}, Llama-3.1-70B-Instruct-RM-RB2 \cite{malik2025rewardbench}, RM-Mistral-7B \cite{raft-2023}, reward-model-Mistral-7B-instruct-Unified-Feedback \cite{regularizing-hidden-states-rm-2024}, Mistral-RM-for-RAFT-GSHF-v0 \cite{raft-2023,iterative-pref-learning-2024}, and pair-preference-model-LLaMA3-8B \cite{rlhf-workflow-2024,slic-hf-2023}
 \item \textbf{DPO-based models} including Nous-Hermes-2-Mistral-7B-DPO \cite{hf-nous-hermes2-mistral7b-dpo} and Matter-0.1-7B-boost-DPO-preview \cite{hf-zerohero-matter01-7b-boost-dpo-preview}.
\end{itemize}
The benchmark consists of 160 pairwise comparisons (Pair), 138 3-way comparisons (Triplet), and 33 4-way comparisons (Quadruplet). For pairwise evaluations, we report accuracy; for multi-way comparisons, we report both full ranking accuracy and best-of-n accuracy.

\subsection{Main Findings}
Table \ref{tab:model_performance} presents the evaluation results of various reward models on DeepResearch RewardBench. From these results, we derive the following key observations: 
\begin{itemize}[leftmargin=1em]
 \item \textbf{Generative RMs outperform traditional sequence classifiers and DPO-based models in the deep research report domain}. This superiority likely stems from the superior generalization and robustness of LLMs, which enable them to perform detailed reasoning to distinguish report quality. In contrast, while traditional reward models perform well on standard benchmarks like RewardBench~\cite{lambert2025rewardbench} and RewardBench2~\cite{malik2025rewardbench}, their lack of specialized training on research reports leads to overall substandard performance in this domain. 

\item \textbf{Ranking remains a considerably more challenging evaluation task than Best-of-N selection}, with difficulty increasing as the number of candidates grows. 
Even leading proprietary models achieve only marginal passing scores on the quadruplet ranking task. This highlights the heightened uncertainty Generative LLMs exhibit when faced with a larger pool of candidate reports—a challenge that stems not only from the inherent complexity of evaluating research reports but also from the rigorous test it imposes on the long-context processing capabilities.

\item \textbf{Performance differences among homogeneous RMs are negligible.} While no RM was intentionally fine-tuned for evaluating Deep Research Reports, there is a clear divide between different RM paradigms, conversely contrasted by the minimal gap within each individual paradigm. This implies that each paradigm may have an inherent ceiling—one where Generative LLMs currently hold a distinct advantage.
\end{itemize}

\subsection{Analysis}
\paragraph{Effect of Report URLs.}

\begin{figure}[t!]
    \centering 
    \includegraphics[width=.95\linewidth]{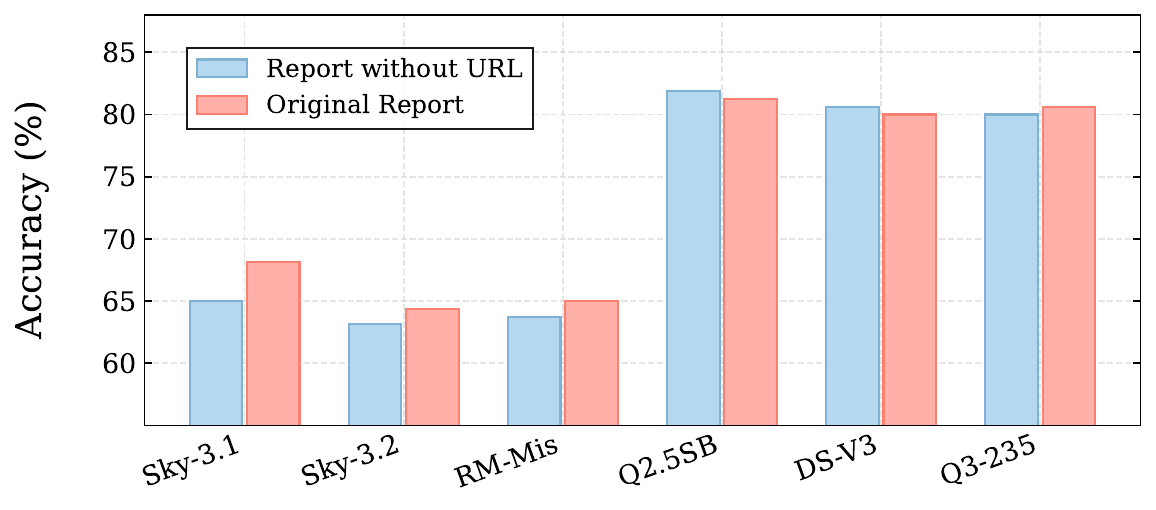}
    \vspace{-0.9em}   
    \caption{Impact of Report URLs on Performance. Left three: Seq Classifiers; Right three: Generative LLMs.}
    \label{fig:urls}
\end{figure}

URLs often account for a non-trivial portion of research reports, substantially extending report length and potentially impacting RMs. To evaluate the stability of these RMs, we deliberately removed the URLs. As illustrated in Figure \ref{fig:urls}, Generative LLMs exhibit exceptional robustness, with their performance remaining virtually unchanged. For sequence classifier RMs, the presence of URLs also exerts a marginal impact, even slightly improving their performance. These findings indicate that URL insertion has little effect on reward models. This is likely because the text-followed-by-URL format is a ubiquitous structural pattern in rich-text formats like Markdown, which both LLMs and RMs encounter extensively during training.

\begin{figure}[t]
    \centering 
    \includegraphics[width=.95\linewidth]{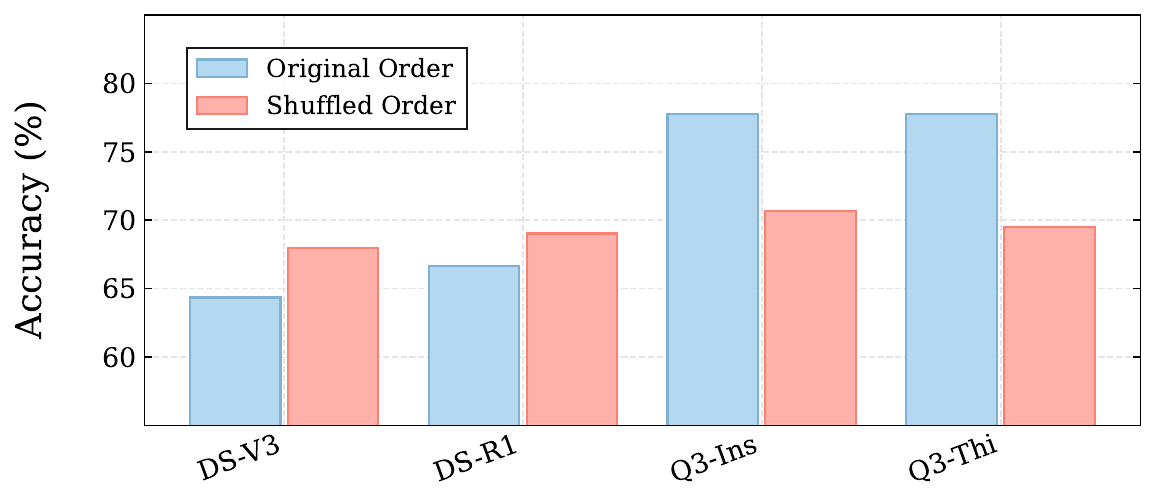}
    \vspace{-1em}   
    \caption{Robustness analysis under input shuffling.}
    \label{fig:shuffle_robustness}
\end{figure}

\paragraph{Effect of Orders.}
For ranking tasks, inputting multiple reports simultaneously is highly likely to introduce positional bias~\citep{pezeshkpour2024large}. We evaluated the robustness of both Thinking and Instruct LLMs against input ordering. As illustrated in Figure \ref{fig:shuffle_robustness}, all evaluated models are significantly affected by the input sequence, with performance fluctuating by as much as 7 percentage points. These results demonstrate that generative LLMs remain highly sensitive to order, highlighting a critical challenge in achieving consistent evaluation for long-form research.

\begin{figure}[t!]
    \centering \includegraphics[width=.9\linewidth]{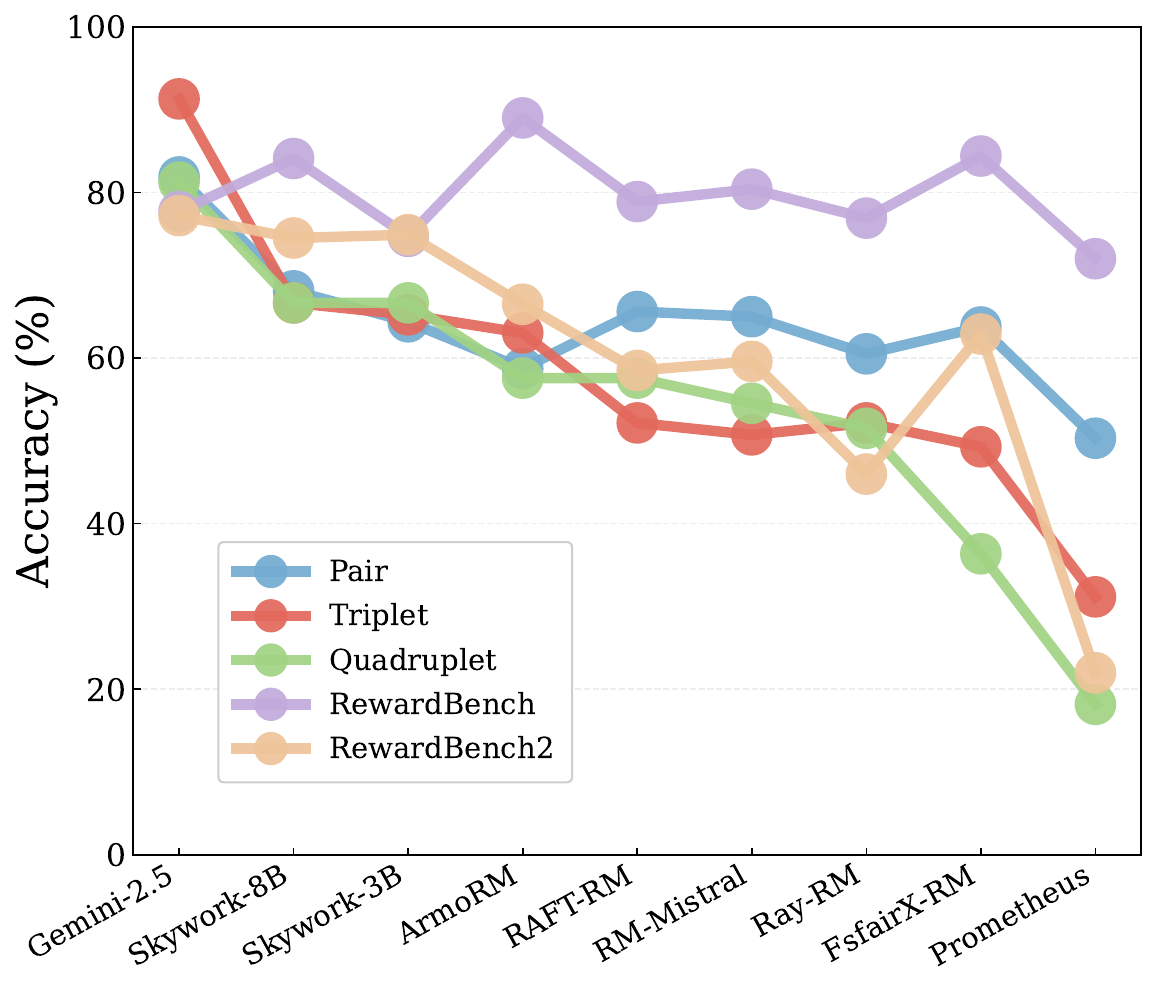}
    \vspace{-0.8em}   
    \caption{Benchmark performance and correlation across RewardBench 1, RewardBench 2, and DeepResearch RewardBench.}
    \label{fig-selected_models}
\end{figure}

\paragraph{Trends Comparison with RewardBench.} Figure \ref{fig-selected_models} illustrates the performance of a diverse range of RMs, varying in both type and scale, evaluated across our Deep Research RewardBench alongside existing mainstream benchmarks (RewardBench 1 and RewardBench 2). A highly consistent trend emerges, demonstrating a strong correlation between research report evaluation capability and general RM performance. Furthermore, it is worth noting that the performance gaps among models on RewardBench 1 are relatively marginal, as most contemporary models have neared saturation in these standard and straightforward scenarios. Conversely, when confronted with more challenging scenarios such as triplet or quadruplet rankings, weaker models exhibit a more pronounced performance degradation compared to binary setups. Overall, an RM's performance on conventional reward modeling benchmarks can serve as an effective proxy for its efficacy in deep research report evaluation scenarios.

\begin{figure*}[ht!]
    \centering 
    \includegraphics[width=.92\linewidth]{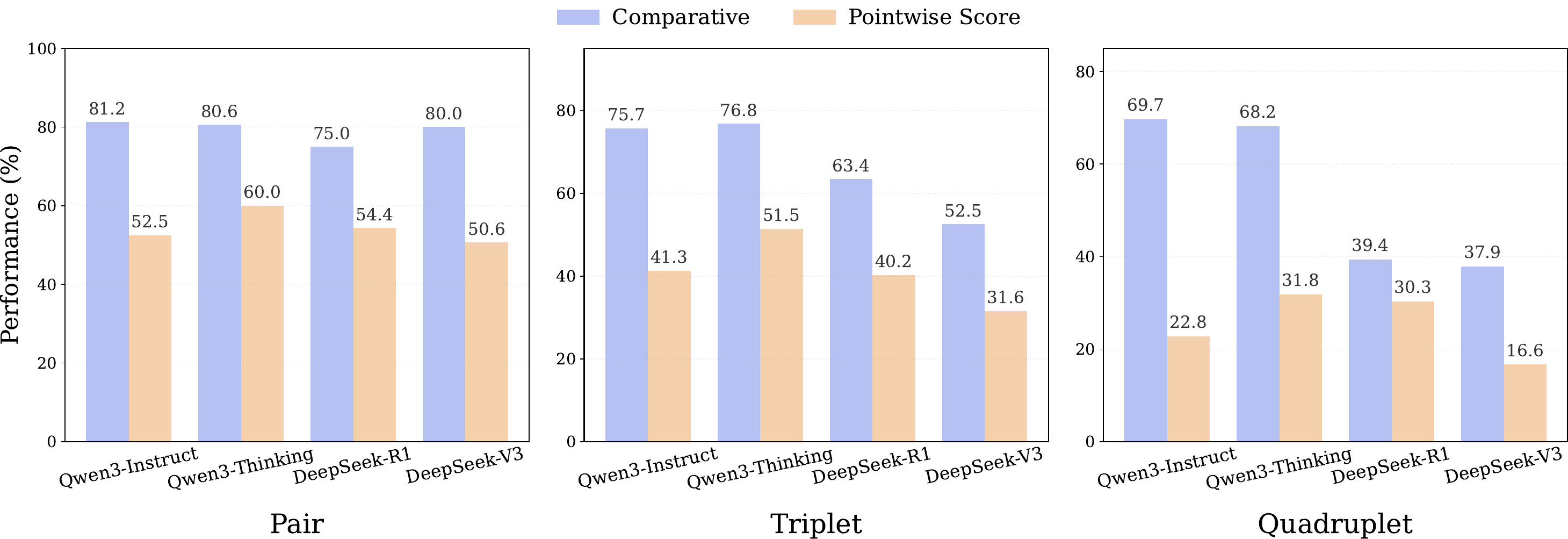}
    \vspace{-0.7em} 
    \caption{Performance of single-point scoring versus comparing multiple reports at once.}
    \label{score-vs-comparsion}
\end{figure*}

\begin{table*}[htbp]
\centering
\small
\renewcommand{\arraystretch}{1.2}
\resizebox{.9\textwidth}{!}{
\begin{tabular}{
    >{\centering\arraybackslash}m{0.9cm}
    >{\centering\arraybackslash}m{0.9cm}
    >{\arraybackslash}m{2.63cm}
    >{\arraybackslash}m{9.0cm}
}

\toprule
\textbf{Abbr.} & \textbf{Ratio} & \textbf{Full Name} & \textbf{General Explanation} \\
\midrule
\textbf{RMI} & 63.5\% & Rubric Misalignment Insensitivity & Judge evaluates based on features orthogonal to the core rubric. \\
\midrule
\textbf{EKE} & 7.70\% & External Knowledge Error & Judge has a personal knowledge gap or factual misunderstanding, mistakenly penalizing true facts as fabricated \\
\midrule
\textbf{CEB} & 7.70\% & \makecell[l]{Citation / Sourcing\\Error Blindness} & Judge overlooks or explicitly tolerates missing or fictional references, often misled by numerical precision \\
\midrule
\textbf{SLB} & 7.70\% &  \makecell[l]{Structural / Length\\Bias} & Judge equates surface organization, length, and visual formatting with deep content quality \\
\bottomrule
\end{tabular}}
\vspace{-0.6em}
\caption{Simplified error taxonomy with concise general explanations. The \textbf{Ratio} column shows each category's share among the manually analyzed wrong judgments; the full distribution under both manual and automated classification, together with representative cases, is reported in Appendix Table~\ref{tab:error-taxonomy-cases}.}
\label{tab:concise-error-taxonomy}
\end{table*}

\paragraph{Comparative Judgment vs. Pointwise Scoring.}
Considering the long-context pressure imposed on LLMs when inputting multiple research reports simultaneously, we also explored an alternative approach: scoring each report independently and ranking them based on the resulting scalar scores. This represents another widely adopted evaluation paradigm in LLMs-as-a-Judge. As illustrated in Figure~\ref{score-vs-comparsion}, comparative judgments consistently outperform pointwise scoring across various models and scenarios, yielding average accuracy gaps of 24.8, 26.0, and 28.4 points for binary, triplet, and quadruplet rankings, respectively. Furthermore, as the candidate pool expands, the accuracy of pointwise scoring exhibits a sharper decline from binary to quadruplet tasks (e.g., Qwen3-Instruct drops from 52.5 to 22.8; DeepSeek-V3 plunges from 50.6 to 16.6).

A granular analysis of specific cases reveals that LLMs tend to cluster their scalar outputs within a narrow range of integer values—most frequently 80 and 85. This heavily causes ties or inverted ranking orders for reports with distinct quality differences. Such numeric insensitivity and the subsequent collapse of the scoring space reflect the inherent difficulty of mapping holistic report quality onto an absolute numerical scale without any anchoring reference. To make scoring truly viable, it demands highly detailed, explicitly verifiable rubrics, paired with the model's capacity to strictly execute them. In a sense, this shifts the challenge of evaluation capability into the realm of complex instruction-following.

\paragraph{Error Taxonomy and Case Study.}
Reasons matter more than scores. To scrutinize why mainstream models underperform, we conducted a manual analysis on 52 failure cases, selected from three widely used models (Qwen3-235B-A22B, DeepSeek-V3-0324, and Gemini-2.5-Flash), where at least two models misjudged the same data point. By comparing the predictions of different models against the golden judge/rubric, we classified four common error categories, as summarized in Table \ref{tab:concise-error-taxonomy}.

As observed, Rubric Misalignment Insensitivity stands out as the most prevalent error, where LLMs evaluate reports using dimensions that deviate from the golden rubric. This finding underscores that the rubric remains the bedrock of research report evaluation, which aligns with the methodology of almost all current studies. Interestingly, other compelling failure modes emerge; for instance, External Knowledge Error occurs when LLMs lack sufficient knowledge reserves (such as up-to-date information) to make a sound judgment, implying that a judge model must possess capabilities equal to or greater than those of the report-generating model. Furthermore, LLMs are frequently ``intimidated” by dense tables or numerical data within the reports, falling prey to such descriptive richness while overlooking whether the data is actually correct or traceable. A more detailed case-by-case breakdown is provided in Table \ref{tab:error-taxonomy-cases}. Notably, when this taxonomy is operationalized as a few-shot LLM classifier and applied to the full set of failure cases, the resulting category distribution closely tracks the manually annotated proportions, suggesting that the taxonomy is amenable to large-scale automated error analysis. We believe this taxonomy will serve as a valuable reference for developing superior evaluation frameworks for deep research reports.
\section{Conclusion}
In this paper, we introduce DeepResearch RewardBench, a pioneering benchmark dedicated to evaluating reward modeling within the context of autonomous deep research reports. We harvest comprehensive reports generated by leading agents and conduct meticulous preference annotation and filtering to guarantee high data quality and a challenging difficulty profile. Our extensive experiments and analysis reveal that while generative LLM-as-a-judge approaches exhibit superior accuracy and robustness compared to traditional reward models, they still suffer from imprecise evaluation dimensions and a limited knowledge base. We expect DeepResearch RewardBench to serve as a vital testbed, fostering a deeper understanding and further optimization of research-oriented agents.


\section*{Limitations}
Although this work represents a pioneering step toward reward modeling evaluation for Deep Research reports, several limitations remain. First, our evaluation primarily focuses on standalone LLMs and RMs. Given the extreme complexity of research report assessment, we believe that integrating agentic frameworks, transitioning toward Agent-as-a-Judge, represents a more promising future direction. Second, the scale of our current dataset is constrained. Due to resource limitations, this study does not extend to a broader range of Deep Research datasets, a gap we intend to bridge in future work.

\section*{Ethical Statement}
We utilized ChatGPT and Grammarly to polish the language of the manuscript. LLMs were employed solely for language refinement and editing purposes, such as improving readability and clarity. They played no role in shaping research ideas, methodology, or analysis. The authors assume full responsibility for all content.

\section*{Acknowledgements} 
This work was supported in part by the Science and Technology Development Fund of Macau SAR (Grant Nos. FDCT/0007/2024/AKP, EF2024-00185-FST), the UM and UMDF (Grant Nos. MYRG-GRG2024-00165-FST-UMDF, MYRG-GRG2025-00236-FST), the Tencent AI Lab Rhino-Bird Research Program (Grant No. EF2023-00151-FST), the Stanley Ho Medical Development Foundation (Grant No. SHMDF-AI/2026/001), and the National Natural Science Foundation of China (Grant No. 62266013).
\bibliography{custom}

@article{du2025deepresearch,
  title={DeepResearch Bench: A Comprehensive Benchmark for Deep Research Agents},
  author={Du, Mingxuan and Xu, Benfeng and Zhu, Chiwei and Wang, Xiaorui and Mao, Zhendong},
  journal={arXiv preprint arXiv:2506.11763},
  year={2025}
}

@article{xu2025researcherbench,
  title={Researcherbench: Evaluating deep ai research systems on the frontiers of scientific inquiry},
  author={Xu, Tianze and Lu, Pengrui and Ye, Lyumanshan and Hu, Xiangkun and Liu, Pengfei},
  journal={arXiv preprint arXiv:2507.16280},
  year={2025}
}

@misc{anthropic2025claudehaiku,
  author = {Anthropic},
  title = {Introducing {Claude} {Haiku} 4.5},
  year = {2025},
  month = {October},
  url = {https://www.anthropic.com/news/claude-haiku-4-5},
  note = {Accessed: 2025-12-30}
}

@article{comanici2025gemini,
  title={Gemini 2.5: Pushing the frontier with advanced reasoning, multimodality, long context, and next generation agentic capabilities},
  author={Comanici, Gheorghe and Bieber, Eric and Schaekermann, Mike and Pasupat, Ice and Sachdeva, Noveen and Dhillon, Inderjit and Blistein, Marcel and Ram, Ori and Zhang, Dan and Rosen, Evan and others},
  journal={arXiv preprint arXiv:2507.06261},
  year={2025}
}

@misc{2410.21276,
  author = {OpenAI},
  title = {GPT-4o System Card},
  eprint = {2410.21276},
  archivePrefix = {arXiv},
  primaryClass = {cs.CL},
  url = {https://arxiv.org/abs/2410.21276},
  doi = {10.48550/arXiv.2410.21276},
  month = {10},
  year = {2024}
}

@article{rafailov2023direct,
  title={Direct preference optimization: Your language model is secretly a reward model},
  author={Rafailov, Rafael and Sharma, Archit and Mitchell, Eric and Manning, Christopher D and Ermon, Stefano and Finn, Chelsea},
  journal={Advances in neural information processing systems},
  volume={36},
  pages={53728--53741},
  year={2023}
}

@inproceedings{gureja2025m,
  title={M-rewardbench: Evaluating reward models in multilingual settings},
  author={Gureja, Srishti and Miranda, Lester James Validad and Islam, Shayekh Bin and Maheshwary, Rishabh and Sharma, Drishti and Winata, Gusti Triandi and Lambert, Nathan and Ruder, Sebastian and Hooker, Sara and Fadaee, Marzieh},
  booktitle={Proceedings of the 63rd Annual Meeting of the Association for Computational Linguistics (Volume 1: Long Papers)},
  pages={43--58},
  year={2025}
}

@article{du2506deepresearch,
  title={Deepresearch bench: A comprehensive benchmark for deep research agents, 2025},
  author={Du, Mingxuan and Xu, Benfeng and Zhu, Chiwei and Wang, Xiaorui and Mao, Zhendong},
  journal={URL https://arxiv. org/abs/2506.11763},
  year={2025}
}

@article{abaskohi2025drbench,
  title = {DRBench: A Realistic Benchmark for Enterprise Deep Research},
  author = {Amirhossein Abaskohi and Tianyi Chen and Miguel Muñoz-Mármol and Curtis Fox and Amrutha Varshini Ramesh and Étienne Marcotte and Xing Han Lù and Nicolas Chapados and Spandana Gella and Issam H. Laradji},
  year = {2025},
  journal = {arXiv (Cornell University)}

}

@article{chen2508browsecomp,
  title = {BrowseComp-Plus: A More Fair and Transparent Evaluation Benchmark of Deep-Research Agent},
  author = {Zijian Chen and Xueguang Ma and Shengyao Zhuang and Ping Nie and Kai Zou and Andrew Liu and Joshua Green and Kshama Patel and Ruoxi Meng and Mingyi Su and Sahel Sharifymoghaddam and Yanxi Li and Haoran Hong and Xinyu Shi and Xuye Liu and Nandan Thakur and Crystina Zhang and Luyu Gao and Wenhu Chen and Jimmy Lin},
  year = {2025},
  journal = {arXiv (Cornell University)}
}

@article{java2025characterizing,
  title={Characterizing deep research: A benchmark and formal definition},
  author={Java, Abhinav and Khandelwal, Ashmit and Midigeshi, Sukruta and Halfaker, Aaron and Deshpande, Amit and Goyal, Navin and Gupta, Ankur and Natarajan, Nagarajan and Sharma, Amit},
  journal={arXiv preprint arXiv:2508.04183},
  year={2025}
}

@article{patel2025deepscholar,
  title={DeepScholar-Bench: A Live Benchmark and Automated Evaluation for Generative Research Synthesis},
  author={Patel, Liana and Arabzadeh, Negar and Gupta, Harshit and Sundar, Ankita and Stoica, Ion and Zaharia, Matei and Guestrin, Carlos},
  journal={arXiv preprint arXiv:2508.20033},
  year={2025}
}

@article{sharma2025researchrubrics,
  title={Researchrubrics: A benchmark of prompts and rubrics for evaluating deep research agents},
  author={Sharma, Manasi and Zhang, Chen Bo Calvin and Bandi, Chaithanya and Wang, Clinton and Aich, Ankit and Nghiem, Huy and Rabbani, Tahseen and Htet, Ye and Jang, Brian and Basu, Sumana and others},
  journal={arXiv preprint arXiv:2511.07685},
  year={2025}
}

@misc{xu2025researcherbenchevaluatingdeepai,
      title={ResearcherBench: Evaluating Deep AI Research Systems on the Frontiers of Scientific Inquiry}, 
      author={Tianze Xu and Pengrui Lu and Lyumanshan Ye and Xiangkun Hu and Pengfei Liu},
      year={2025},
      eprint={2507.16280},
      archivePrefix={arXiv},
      primaryClass={cs.AI},
      url={https://arxiv.org/abs/2507.16280}, 
}

@article{wan2025deepresearch,
  title={DeepResearch Arena: The First Exam of LLMs' Research Abilities via Seminar-Grounded Tasks},
  author={Wan, Haiyuan and Yang, Chen and Yu, Junchi and Tu, Meiqi and Lu, Jiaxuan and Yu, Di and Cao, Jianbao and Gao, Ben and Xie, Jiaqing and Wang, Aoran and others},
  journal={arXiv preprint arXiv:2509.01396},
  year={2025}
}

@article{liang2025towards,
  title={Towards Personalized Deep Research: Benchmarks and Evaluations},
  author={Liang, Yuan and Li, Jiaxian and Wang, Yuqing and Wang, Piaohong and Tian, Motong and Liu, Pai and Qiao, Shuofei and Fang, Runnan and Zhu, He and Zhang, Ge and others},
  journal={arXiv preprint arXiv:2509.25106},
  year={2025}
}

@article{gou2025mind2web,
  title={Mind2web 2: Evaluating agentic search with agent-as-a-judge, 2025},
  author={Gou, Boyu and Huang, Zanming and Ning, Yuting and Gu, Yu and Lin, Michael and Qi, Weijian and Kopanev, Andrei and Yu, Botao and Guti{\'e}rrez, Bernal Jim{\'e}nez and Shu, Yiheng and others},
  journal={URL https://arxiv. org/abs/2506.21506},
  year={2025}
}

@inproceedings{lambert2025rewardbench,
  title={Rewardbench: Evaluating reward models for language modeling},
  author={Lambert, Nathan and Pyatkin, Valentina and Morrison, Jacob and Miranda, Lester James Validad and Lin, Bill Yuchen and Chandu, Khyathi and Dziri, Nouha and Kumar, Sachin and Zick, Tom and Choi, Yejin and others},
  booktitle={Findings of the Association for Computational Linguistics: NAACL 2025},
  pages={1755--1797},
  year={2025}
}

@article{malik2025rewardbench,
  title={RewardBench 2: Advancing Reward Model Evaluation},
  author={Malik, Saumya and Pyatkin, Valentina and Land, Sander and Morrison, Jacob and Smith, Noah A and Hajishirzi, Hannaneh and Lambert, Nathan},
  journal={arXiv preprint arXiv:2506.01937},
  year={2025}
}

@article{dai2023safe,
  title={Safe rlhf: Safe reinforcement learning from human feedback},
  author={Dai, Josef and Pan, Xuehai and Sun, Ruiyang and Ji, Jiaming and Xu, Xinbo and Liu, Mickel and Wang, Yizhou and Yang, Yaodong},
  journal={arXiv preprint arXiv:2310.12773},
  year={2023}
}

@article{chaudhari2025rlhf,
  title={Rlhf deciphered: A critical analysis of reinforcement learning from human feedback for llms},
  author={Chaudhari, Shreyas and Aggarwal, Pranjal and Murahari, Vishvak and Rajpurohit, Tanmay and Kalyan, Ashwin and Narasimhan, Karthik and Deshpande, Ameet and Castro da Silva, Bruno},
  journal={ACM Computing Surveys},
  volume={58},
  number={2},
  pages={1--37},
  year={2025},
  publisher={ACM New York, NY}
}

@misc{qwen3-techreport,
  title        = {Qwen3 Technical Report},
  author       = {Qwen Team},
  year         = {2025},
  eprint       = {2505.09388},
  archivePrefix= {arXiv},
  primaryClass = {cs.CL}
}

@misc{deepseek-r1-2025,
  title        = {DeepSeek-R1: Incentivizing Reasoning Capability in LLMs via Reinforcement Learning},
  author       = {DeepSeek-AI},
  year         = {2025},
  eprint       = {2501.12948},
  archivePrefix= {arXiv}
}

@misc{deepseek-v3-techreport-2024,
  title        = {DeepSeek-V3 Technical Report},
  author       = {DeepSeek-AI},
  year         = {2024},
  eprint       = {2412.19437},
  archivePrefix= {arXiv}
}

@misc{kimi-k2-techreport-2025,
  title  = {Kimi-K2 Technical Report},
  author = {Moonshot AI},
  year   = {2025}
}

@misc{prometheus2-2024,
  title  = {Prometheus 2: An Open Source Language Model Specialized in Evaluating Other Language Models},
  author = {Seungone Kim and Juyoung Suk and Shayne Longpre and others},
  year   = {2024},
  eprint = {2405.01535}
}

@article{skywork-reward-v2-2025,
  title   = {Skywork-Reward-V2: Scaling Preference Data Curation via Human-AI Synergy},
  author  = {Liu, Chris Yuhao and Zeng, Liang and others},
  journal = {arXiv preprint arXiv:2507.01352},
  year    = {2025}
}

@inproceedings{armorm-emnlp-2024,
  title     = {Interpretable Preferences via Multi-Objective Reward Modeling and Mixture-of-Experts},
  author    = {Haoxiang Wang and Wei Xiong and Tengyang Xie and Han Zhao and Tong Zhang},
  booktitle = {EMNLP},
  year      = {2024}
}

@article{raft-2023,
  title   = {Raft: Reward ranked finetuning for generative foundation model alignment},
  author  = {Dong, Hanze and Xiong, Wei and others},
  journal = {arXiv preprint arXiv:2304.06767},
  year    = {2023}
}

@misc{iterative-pref-learning-2024,
  title  = {Iterative Preference Learning from Human Feedback: Bridging Theory and Practice for RLHF under KL-Constraint},
  author = {Wei Xiong and Hanze Dong and others},
  year   = {2024},
  eprint = {2312.11456}
}

@article{regularizing-hidden-states-rm-2024,
  title   = {Regularizing Hidden States Enables Learning Generalizable Reward Model for LLMs},
  author  = {Yang, Rui and Ding, Ruomeng and others},
  journal = {arXiv preprint arXiv:2406.10216},
  year    = {2024}
}

@misc{rlhf-workflow-2024,
  title  = {RLHF Workflow: From Reward Modeling to Online RLHF},
  author = {Hanze Dong and Wei Xiong and others},
  year   = {2024},
  eprint = {2405.07863}
}

@misc{slic-hf-2023,
  title  = {Slic-hf: Sequence likelihood calibration with human feedback},
  author = {Zhao, Yao and Joshi, Rishabh and others},
  year   = {2023},
  eprint = {2305.10425}
}

@misc{openai-gpt41-docs,
  author       = {OpenAI},
  title        = {GPT-4.1 Model Overview},
  year         = {2026},
  url          = {https://developers.openai.com/api/docs/models#gpt-4-1},
  urldate      = {2026-08-31},
  note         = {Product documentation}
}

@misc{google-gemini25-flash-docs,
  author       = {Google DeepMind},
  title        = {Gemini API Models: Gemini 2.5 Flash},
  year         = {2026},
  url          = {https://ai.google.dev/gemini-api/docs/models},
  urldate      = {2026-08-31},
  note         = {Product documentation}
}

@misc{hf-nous-hermes2-mistral7b-dpo,
  author       = {Teknium and theemozilla and karan4d and {huemin\_art}},
  title        = {Nous Hermes 2 - Mistral 7B - DPO},
  year         = {2024},
  url          = {https://huggingface.co/NousResearch/Nous-Hermes-2-Mistral-7B-DPO},
  urldate      = {2026-08-31},
  note         = {Hugging Face model card}
}

@misc{hf-zerohero-matter01-7b-boost-dpo-preview,
  author       = {0-hero},
  title        = {Matter 7B - 0.1 Boost - DPO (Mistral 7B Finetune)},
  year         = {2024},
  url          = {https://huggingface.co/0-hero/Matter-0.1-7B-boost-DPO-preview},
  urldate      = {2026-08-31},
  note         = {Hugging Face model card}
}

@inproceedings{pezeshkpour2024large,
  title={Large language models sensitivity to the order of options in multiple-choice questions},
  author={Pezeshkpour, Pouya and Hruschka, Estevam},
  booktitle={Findings of the Association for Computational Linguistics: NAACL 2024},
  pages={2006--2017},
  year={2024}
}

@inproceedings{li2025vlrewardbench,
  title={VL-RewardBench: A Challenging Benchmark for Vision-Language Generative Reward Models},
  author={Li, Lei and Wei, Yuancheng and Xie, Zhihui and Yang, Xuqing and Song, Yifan and Wang, Peiyi and An, Chenxin and Liu, Tianyu and Li, Sujian and Lin, Bill Yuchen and Kong, Lingpeng and Liu, Qi},
  booktitle={{CVPR}},
  year={2025}
}

@inproceedings{liu2025rm,
  title={Rm-bench: Benchmarking reward models of language models with subtlety and style},
  author={Liu, Yantao and Yao, Zijun and Min, Rui and Cao, Yixin and Hou, Lei and Li, Juanzi},
  booktitle={International Conference on Learning Representations},
  volume={2025},
  pages={44323--44355},
  year={2025}
}

@article{liu2025compassverifier,
  title={CompassVerifier: A Unified and Robust Verifier for LLMs Evaluation and Outcome Reward},
  author={Liu, Shudong and Liu, Hongwei and Liu, Junnan and Xiao, Linchen and Gao, Songyang and Lyu, Chengqi and Gu, Yuzhe and Zhang, Wenwei and Wong, Derek F and Zhang, Songyang and Chen, Kai},
  journal={arXiv preprint arXiv:2508.03686},
  year={2025}
}

@article{zhong2025comprehensive,
  title={A comprehensive survey of reward models: Taxonomy, applications, challenges, and future},
  author={Zhong, Jialun and Shen, Wei and Li, Yanzeng and Gao, Songyang and Lu, Hua and Chen, Yicheng and Zhang, Yang and Zhou, Wei and Gu, Jinjie and Zou, Lei},
  journal={arXiv preprint arXiv:2504.12328},
  year={2025}
}

\clearpage
\appendix
\label{sec:appendix}

{\color{black}
\section{Evaluation Rubrics and Scoring Protocols}
\label{appendix_rubrics}

Our benchmark adopts the native evaluation protocols from two source datasets. We describe each protocol below and provide representative examples to illustrate how reports are scored.

\subsection{DeepResearch Bench: The RACE Framework}
\label{appendix_race}

DeepResearch Bench~\cite{du2025deepresearch} proposes the \textbf{RACE} (Reference-based Adaptive Criteria-driven Evaluation) framework for report quality assessment. RACE evaluates reports along four top-level orthogonal dimensions, as defined in Table~\ref{tab:race_dimensions}.

\begin{table}[h!]
\centering
\resizebox{0.46\textwidth}{!}{
\begin{tabular}{lp{5.5cm}}
\toprule
\textbf{Dimension} & \textbf{Description} \\ \midrule
Comprehensiveness (COMP) & Report covers key areas of the topic, ensures overall understanding, and does not omit important parts. \\
Insight/Depth (DEPTH) & Report deeply analyzes causes, impacts, and trends, providing valuable insights. \\
Instruction-Following (INST) & Report closely follows the research topic and directly answers questions. \\
Readability (READ) & Report has a clear structure, fluent language, and is easy to understand. \\ \bottomrule
\end{tabular}}
\caption{Definitions of the four RACE evaluation dimensions, as established in DeepResearch Bench~\cite{du2025deepresearch}.}
\label{tab:race_dimensions}
\end{table}

Given a research task, the Judge LLM first generates task-specific weights $W_d$ for each dimension $d$ by averaging weights over $T$ independent trials. Then, for each dimension, the Judge LLM generates a set of fine-grained criteria $\{c_{d,k}\}$ with corresponding criterion weights $\{w_{d,k}\}$. A high-quality reference report $R_{\mathrm{ref}}$ is selected, and both the target report and reference report are scored against all criteria. The final score is computed as a relative score:
$$S_{\mathrm{final}}(R_{\mathrm{tgt}}) = \frac{S_{\mathrm{int}}(R_{\mathrm{tgt}})}{S_{\mathrm{int}}(R_{\mathrm{tgt}}) + S_{\mathrm{int}}(R_{\mathrm{ref}})}$$
where $S_{\mathrm{int}}(R)$ is the weighted combination of dimension-level scores.

\paragraph{Example.} For the research task ``Researching how the world's wealthiest governments invest'' , the RACE framework produces sub-scores for each dimension. Report A receives: Comprehensiveness 53.56, Insight 58.24, Instruction Following 52.77, Readability 55.05, Overall 55.34. Report B receives: Comprehensiveness 45.23, Insight 51.12, Instruction Following 46.89, Readability 49.68, Overall 47.98. The multi-dimensional scoring reveals that Report A's advantage is most pronounced in Insight/Depth (+7.12) rather than a uniform improvement across all dimensions.

\subsection{ResearcherBench: Rubric Assessment}
\label{appendix_researcherbench}

ResearcherBench~\cite{xu2025researcherbench} employs a fine-grained \textbf{rubric assessment} framework where each research question is decomposed into multiple expert-designed rubric items. Each rubric item represents a specific aspect of reasoning or insight that should be addressed, and is assigned an importance weight by domain experts.

The evaluation process consists of three steps:
\begin{enumerate}[leftmargin=1.5em, topsep=2pt]
    \item \textbf{Insight Extraction}: Key insights are extracted from contextual sources (discussion records, literature, expert opinions) using Claude-3.7-Sonnet, then validated by domain experts.
    \item \textbf{Rubric Item Design}: Human annotators (Master's/PhD students in AI) transform extracted insights into operational rubric items with importance weights.
    \item \textbf{Quality Control}: Each rubric is collaboratively developed by two AI researchers, with review, pilot testing, and iterative refinement.
\end{enumerate}

The coverage score for a question $Q_k$ is computed as:
$$\text{Coverage Score} = \frac{\sum_{i=1}^{n} w_i \cdot c_i}{\sum_{i=1}^{n} w_i}$$
where $w_i$ is the weight of the $i$-th rubric item and $c_i \in \{0, 1\}$ is the binary judgment by o3-mini on whether the response covers that rubric item.

\paragraph{Example.} For a question on improving LLM effectiveness on long text reasoning tasks, expert-designed rubric items might include: 1) Discussion of model architecture modifications for long-context (weight 0.3), 2) Training strategies for extended context (weight 0.25), 3) Inference-time strategies such as retrieval augmentation (weight 0.25), 4) Evaluation methods for long-context capabilities (weight 0.2). If a DARS response covers items 1, 2, and 4 but misses item 3, the coverage score would be $\frac{0.3 \times 1 + 0.25 \times 1 + 0.25 \times 0 + 0.2 \times 1}{0.3 + 0.25 + 0.25 + 0.2} = 0.75$.

A real benchmark instance with full o3-mini coverage judgments and justifications for two candidate reports is shown in Table~\ref{tab:rb_rubric_full}.


Table~\ref{tab:rb_rubric_full} reveals a near-mirror-image coverage pattern: the chosen report (gpt-4o-search-preview) covers the distillation-centric items (\#2, \#6, \#9) that articulate why distillation outperforms RL on small models, whereas the rejected report (Perplexity) captures only the complementary direct-RL evidence item (\#1). Both nevertheless miss the same three specific items -- multi-stage pipeline (\#3), self-improvement RL (\#4), and industry compression case studies (\#7) -- so even the chosen report is far from complete. The 0.556 vs.\ 0.333 gap therefore traces to which side of the RL-vs.-distillation debate each report emphasizes, amplified by the rubric placing its heaviest weight ($w_2 = 3$) on the ``why-distillation-wins'' item. Despite this clear-cut rubric verdict in favor of A, all three generative judges in our error analysis still side with B, making this case the canonical Rubric Misalignment Insensitivity (RMI) failure revisited in Appendix~\ref{sec:appendix-failure-modes}.

\subsection{Mapping to Preference Labels}

In our benchmark, the scores produced by RACE (for DeepResearch Bench tasks) and the coverage scores (for ResearcherBench tasks) serve as the basis for constructing preference tuples. Reports within each tuple are ranked by their scores, with the highest-scoring report designated as \emph{Chosen} and the others as \emph{Rejected}. The score margin (defined as the absolute difference between report scores on a 100-point scale) is used to control task difficulty: we exclude pairs with margins exceeding 10 points to ensure discriminative challenge. This process is described in detail in Section 3.
}

\section{Annotation Details}
\label{Annotation Details}
For each research task, we have three Ph.D. annotators independently assigned a ranking. Any inconsistencies in the rankings prompted a meta-evaluation, where the annotators discussed the cases to reach a consensus ranking. If a consensus still could not be reached after discussion, the data point was discarded from the dataset, as different responses might each possess unique advantages without a clear winner. The initial inter-annotator agreement stood at 63\%, which improved to approximately 85\% after the discussion phase.

{\color{black}
A potential concern is circularity: since our ground-truth preference labels are derived from LLM-as-a-Judge (Gemini-2.5-Pro for DeepResearch Bench and o3-mini for ResearcherBench), evaluating other generative LLMs on these labels could be self-fulfilling. We address this concern as follows. First, among all 19 evaluated models in Table~\ref{tab:model_performance}, only GPT-4.1 partially overlaps with the label generation pipeline---it was used in an early stage of scoring for a subset of ResearcherBench instances, subject to subsequent human quality control. The two primary Judge LLMs (Gemini-2.5-Pro and o3-mini) are \emph{not} among the evaluated models. Second, all labels underwent human verification, ensuring the ground truth is not purely LLM-generated. Third, sequence classifiers---which share no architecture or training data with the Judge LLMs---also show consistent relative rankings across subsets.
}

\section{Example: Deep Research Bench Case \#52}

\onecolumn
\begin{table*}[t]
\centering
\footnotesize
\resizebox{\textwidth}{!}{
\begin{tabular}{clp{3.8cm}cp{4.4cm}cp{4.4cm}}
\toprule
\textbf{\#} & \textbf{$w_i$} & \textbf{Rubric item} & \multicolumn{2}{c}{\textbf{Report A: gpt-4o-search-preview}} & \multicolumn{2}{c}{\textbf{Report B: Perplexity}} \\
\cmidrule(lr){4-5} \cmidrule(lr){6-7}
 &  &  & $c_i$ & \textbf{o3-mini justification (abbrev.)} & $c_i$ & \textbf{o3-mini justification (abbrev.)} \\
\midrule
1 & 2 & Direct application of RL to small ($\sim$7B) models produces substantial gains, with experimental evidence.
  & $\times$ & Mainly theoretical discussion on RL-vs.-distillation challenges; no experimental evidence or measured improvements on 7B-scale models.
  & \checkmark & Includes detailed experimental evidence of substantial performance improvements from applying RL directly to smaller models. \\[2pt]
2 & 3 & Why distillation from larger models is more effective than RL for enhancing small models' capabilities.
  & \checkmark & Explains RL is less practical for small models (compute constraints, sample inefficiency, overfitting); contrasts with distillation as more efficient transfer.
  & $\times$ & Provides balanced RL-vs.-distillation comparison but does not explicitly explain why distillation is superior. \\[2pt]
3 & 2 & A multi-stage training process (base prep $\to$ RL in verifiable domains $\to$ iterative SFT+RL) for small models.
  & $\times$ & Discusses RL and distillation generally; no mention of base $\to$ RL-in-math/code $\to$ iterative SFT+RL pipeline.
  & $\times$ & Covers various aspects of applying RL to small models but does not describe the specified multi-stage pipeline. \\[2pt]
4 & 2 & Self-improvement RL enables small models to extend reasoning beyond initial token-length limits.
  & $\times$ & Discusses RL / distillation in general but does not address self-improvement RL or reasoning-chain extension.
  & $\times$ & Discusses RL for small models vs.\ distillation; does not address self-improvement RL or token-length extension. \\[2pt]
5 & 2 & Parameter-capacity limits of small models hinder RL for complex reasoning.
  & \checkmark & Explicitly explains that limited capacity (overfitting risks, sample inefficiency) hinders RL for complex reasoning.
  & \checkmark & Mentions ``squashing'' of rare behaviors in small models and contrasts with RL benefits, implicitly capturing capacity limits. \\[2pt]
6 & 2 & Transferring pre-optimized reasoning patterns via distillation is more effective than learning new RL optimizations.
  & \checkmark & Contrasts RL challenges (compute, sample efficiency, overfitting) with distillation's pre-optimized transfer advantage.
  & $\times$ & Notes complementary nature of RL and distillation but does not state distillation's pre-optimization superiority. \\[2pt]
7 & 2 & Industry case studies of distillation achieving superior model compression vs.\ direct RL.
  & $\times$ & Compares RL and distillation, cites DistilBERT, but no explicit industry case studies on compression.
  & $\times$ & Details research findings and RL/distillation comparisons but no industry case studies on compression in current practice. \\[2pt]
8 & 2 & How model size shapes the relative effectiveness of RL vs.\ distillation.
  & \checkmark & Explains small-model RL challenges and contrasts with distillation as effective transfer from large models.
  & \checkmark & Explains tailored RL strategies for small models (smaller batches, different optimization) and distillation's stability; directly addresses size effect. \\[2pt]
9 & 1 & Distillation circumvents the compute / instability risks of directly applying RL to small models.
  & \checkmark & Explicitly states RL's compute constraints and instability (overfitting); contrasts with distillation's resource-efficient transfer.
  & $\times$ & Briefly mentions distillation's stability vs.\ RL but does not explicitly link to compute and instability risks. \\
\midrule
\multicolumn{3}{r}{\textbf{Weighted-covered $\sum_i w_i c_i$ (of $\sum_i w_i = 18$)}}
  & \multicolumn{2}{c}{\textbf{10}} & \multicolumn{2}{c}{\textbf{6}} \\
\multicolumn{3}{r}{\textbf{Coverage / Recall score}}
  & \multicolumn{2}{c}{\textbf{0.556}} & \multicolumn{2}{c}{\textbf{0.333}} \\
\bottomrule
\end{tabular}}
\caption{Full ResearcherBench rubric evaluation for DRRB id=120 (ResearcherBench id=25, ``RL techniques for small LLMs vs.\ distillation''), as scored by o3-mini. $c_i \in \{0, 1\}$ denotes whether each rubric item is covered by the report; justifications are condensed from o3-mini's verbatim outputs. The two recall scores (Report A: 0.556 vs.\ Report B: 0.333, equivalently 55.6 vs.\ 33.3 on the 100-point scale) form the basis for the preference label of this pair.}
\label{tab:rb_rubric_full}
\end{table*}

\clearpage
\onecolumn
\begin{tcolorbox}[colback=white,colframe=black,width=\textwidth,
title={\small\textbf{Example Case 52: Sovereign Wealth Fund Research}},
fonttitle=\small,fontlower=\tiny,
left=2mm,right=2mm,top=2mm,bottom=2mm]

\textcolor{blue}{\textbf{Question}} Researching how the world's wealthiest governments invest.

\vspace{0.3em}

\textcolor{blue}{\textbf{Report A Score}} 55.34 (Rank 1)

\textcolor{blue}{\textbf{Report A Response}} 

\begin{quote}\tiny
How the World's Wealthiest Governments Invest: Mandates, Governance, Asset Mixes, Performance, and Strategic Shifts (2005--2025)

Executive summary

The world's wealthiest governments concentrate public capital in two main forms: sovereign wealth funds (SWFs) and official foreign-exchange reserve funds. To provide a rigorous, comparable view, two definitions of wealthiest governments are used in parallel. Definition A ranks by SWF assets under management (AUM). Definition B ranks by official reserve assets and, where relevant, net international investment position (NIIP). Under Definition A, Norway's Government Pension Fund Global (GPFG), Abu Dhabi's ADIA, Singapore's GIC and Temasek, Kuwait's KIA, Qatar's QIA, Saudi Arabia's PIF, Australia's Future Fund, Hong Kong's Exchange Fund, Russia's National Wealth Fund (NWF), and Korea Investment Corporation (KIC) dominate global sovereign investment activity...

Performance varies by mandate. GPFG returned 13.1\% in 2024 and reported a 1998--2024 average annual return of 6.3\% with a 4.1\% average annual net real return; its end-2024 asset mix was roughly 71\% equities, 27\% fixed income and 2\% unlisted real estate/renewables, and AUM stood at NOK 19,742 billion (about USD 1.8--1.9 trillion at prevailing FX). GIC's 20-year nominal USD return to 31 March 2025 was 5.7\% and the real return 3.8\%, with a diversified 51/26/23 mix across equities, fixed income and real assets. Temasek's Net Portfolio Value reached S\$434 billion at 31 March 2025 (US\$324 billion), with 20-year total shareholder return of 7\% and one-year S\$ TSR of 11.8\%...

[Full report continues with detailed sections on fund-by-fund profiles, comparative analysis, methods, and policy implications]
\end{quote}

\vspace{0.2em}
\textcolor{blue}{\textbf{Meta Data A}}
\begin{itemize}\setlength\itemsep{0em}
    \item Model: thinkdepthai-deepresearch
    \item Overall Score: 55.34
    \item Rank: 1
    \item Comprehensiveness Score: 53.56
    \item Insight Score: 58.24
    \item Instruction Following Score: 52.77
    \item Readability Score: 55.05
    \item Original Dataset Question ID: 53
\end{itemize}

\vspace{0.5em}

\textcolor{blue}{\textbf{Report B Score}} 47.98 (Rank 2)

\textcolor{blue}{\textbf{Report B Response}}

\begin{quote}\tiny
Global State Capital: Investment Strategies of the World's Wealthiest Governments

Executive Summary

The management and investment of national wealth by the world's most affluent governments represent a significant force in the global financial landscape. Identifying these ``wealthiest'' governments is complex, depending on metrics such as Gross Domestic Product (GDP), total national wealth, or the size of state-controlled investment vehicles like Sovereign Wealth Funds (SWFs). The United States and China consistently lead in nominal GDP and total national wealth, while resource-rich nations like Norway and Gulf states, alongside financial centers like Singapore, command attention through their massive SWFs...

Investment strategies vary significantly, driven by diverse mandates ranging from long-term intergenerational savings and economic stabilization to strategic development and economic diversification. While traditional asset classes like public equities and fixed income remain core holdings, there is a clear and growing trend towards alternative assets, including private equity, infrastructure, real estate, and direct strategic investments...

[Full report includes 8 sections covering wealth metrics, SWF rankings, investment strategies, governance frameworks, fund profiles, and future trends]
\end{quote}

\vspace{0.2em}
\textcolor{blue}{\textbf{Meta Data B}}
\begin{itemize}\setlength\itemsep{0em}
    \item Model: gemini-2.5-pro-deepresearch
    \item Overall Score: 47.98
    \item Rank: 2
    \item Comprehensiveness Score: 45.23
    \item Insight Score: 51.12
    \item Instruction Following Score: 46.89
    \item Readability Score: 49.68
    \item Original Dataset Question ID: 53
\end{itemize}



\end{tcolorbox}
\clearpage
\twocolumn

\section{Failure Mode Anatomy: Representative Cases and Automated Validation of the Taxonomy}
\label{sec:appendix-failure-modes}

\paragraph{Representative Cases.}
To make these failure modes concrete, we examine one canonical case per main category; full reports, verbatim CoT excerpts, and per-case diagnoses are deferred to Table~\ref{tab:error-taxonomy-cases}.
For \textbf{RMI} (case id=120, RL-for-small-LLMs), all three judges reward the rejected report for citation freshness and concrete RL benchmarks (e.g., AMC23 63$\to$80\%), while the rubric's heaviest weight (w=6/18) is on \emph{why distillation outperforms RL}---three sub-points entirely covered by the chosen report and entirely missed by the rejected one. The judge's stated criteria simply do not appear in the rubric.
For \textbf{EKE} (case id=40, top-grossing Chinese films), both reports verbatim cite ``154.46 yi yuan'' ($\approx$\$2.1B, the real Nezha~2 record from January 2025), yet Qwen3-235B and Gemini flag this as fabricated by misreading the Chinese unit \emph{yi} ($10^{8}$) as \emph{billion} ($10^{9}$); strikingly, the same figure appears in the rejected report without being flagged, confirming that the error lives in judge knowledge, not in report content.
For \textbf{CEB} (case id=139, lifelong learning), the chosen report contains \emph{zero} references yet produces specific-looking artifacts (``60.82\% power reduction'', ``Nature Communications 2025''); Gemini \emph{explicitly} acknowledges that the citations may be fictional, yet still rewards the report for ``specificity''---numerical precision is read as evidence regardless of sourcing.
For \textbf{SLB} (case id=19, Anthropic Streamable HTTP), the rejected report is concise (6.2~KB) and cites the canonical MCP specification, while the chosen report is 4$\times$ longer with diagrams, code blocks, section headings, and a numeric benchmark drawn from a single non-authoritative blog; DeepSeek-V3's deciding factors are entirely structural.
Across these cases, judges fail not because they miss content but because they implicitly evaluate axes \emph{different} from the rubric---freshness, surface specificity, structural richness, or world-model plausibility---rather than rubric-defined reference-point coverage.

\paragraph{Cross-cutting Patterns.}
Two complementary lenses---a fine-grained manual annotation over a curated subset of hard cases, and an LLM-based classification applied to the full set of failure cases---yield consistent observations regarding the relative prevalence of each failure mode. Rubric Misalignment Insensitivity (RMI) dominates by a wide margin in both views, capturing well over half of all judge errors, while the remaining categories (EKE, CEB, SLB, and the residual ``other'' bucket) each occupy a single-digit-to-low-teens share with no notable rank flips between the two analyses. Beyond aggregate prevalence, judge-specific concentrations also persist across the two views: EKE failures cluster on Qwen3-235B, surface-bias (SLB) failures cluster on DeepSeek-V3, and the empty-CoT pathology appears \emph{exclusively} on Gemini-2.5-Flash. CEB, in contrast, is spread approximately uniformly across all three judges, and RMI is essentially flat---every judge family produces RMI at roughly the same rate. This contrast is informative: judge-specific failures reflect model-family weaknesses (knowledge cutoffs, surface-feature biases, prompt-following lapses) that ensembling across families can in principle mitigate, whereas the dominant RMI mode reflects a \emph{structural} gap between how LLM judges evaluate and what reference-based rubrics actually score---it persists regardless of which judge is used.

\paragraph{Automatability of the Taxonomy.}
To assess whether the proposed taxonomy can support large-scale failure-mode analysis without continued manual labor, we operationalize it as a few-shot LLM classifier and apply it to every wrong judgment produced by the three generative judges. The classifier is prompted with the category definitions, the decision flow (verify the judge's CoT claims against the actual report; map to category A if they fail, to RMI if they hold, to EKE if the failure traces to the judge's own world knowledge), and four anchor examples drawn from our manual annotation. We then validate the automated labels against the manually annotated subset and find an exact-match agreement rate of approximately 87\%, rising to roughly 90\% when restricted to high-confidence manual labels; nearly all remaining discrepancies trace to a single boundary---numerical-precision-driven attributions that can be plausibly read as either CEB (overlooked sourcing) or RMI (sourcing axis simply not in the rubric). The headline category distribution is reproduced almost exactly: RMI remains the dominant failure mode at roughly 62\% under automated classification versus 63.5\% under manual annotation, while EKE, CEB, and SLB each settle in the single-digit-to-low-teens range with no rank flips between the two views. Judge-specific concentration patterns are likewise preserved---over 80\% of EKE failures concentrate on Qwen3-235B, around 80\% of SLB failures on DeepSeek-V3, and 100\% of empty-CoT failures on Gemini-2.5-Flash---indicating that the taxonomy is sufficiently well-defined and the decision rules sufficiently explicit to be applied by an LLM at scale. As an additional sanity check, the automated classification surfaces a clean dataset-level signature that the manual subset hinted at but could not fully establish: on the Chinese-language deep-research subset, structural and breadth-vs-coverage failures (SLB, DBM) jointly account for around 19\% of all judge errors, whereas they are virtually absent (under 2\%) on the English-language scientific-research subset; conversely, knowledge-gap and sourcing failures (EKE, CEB) jointly account for over 30\% on the scientific-research subset against under 10\% on the deep-research subset. This domain-dependent split is intrinsic to the full-set analysis rather than an artifact of selection bias, and suggests that benchmark coverage decisions---which content domains to include---will shape which failure modes a given evaluation framework surfaces.

\paragraph{Implications.}
These observations point to two complementary directions for improving deep-research evaluation. First, judge ensembling across model families is necessary to dilute the judge-specific failures (EKE on Qwen3, SLB on DeepSeek, empty-CoT on Gemini), since no single judge family is free of model-specific knowledge gaps and surface biases. Second---and more fundamentally---no amount of judge ensembling resolves RMI, which accounts for the dominant majority of failures in both the manual and the automated analyses. Closing the rubric--judge gap likely requires either (i) anchoring judge prompts directly to the rubric's reference points so that the judge's CoT scoring axis aligns with the rubric's scoring axis, or (ii) sidestepping absolute rubric scoring altogether via comparative judgment, as in the DRRB protocol, where the judge produces a single relative preference rather than projecting reports onto its own implicit quality axes. Our analysis suggests that scaling up judge capability alone is unlikely to close the gap.

\clearpage
\onecolumn
\begin{table*}[t]
\centering
\renewcommand{\arraystretch}{1.0}
\setlength{\tabcolsep}{3pt}
\footnotesize

\newcolumntype{L}[1]{>{\raggedright\arraybackslash}p{#1}}

\begin{tabular}{@{}L{1.15cm}L{1.0cm}L{1.0cm}L{2.1cm}L{3.15cm}L{3.15cm}L{3.15cm}@{}}
\toprule
\textbf{Category \& case} & \textbf{Manual} & \textbf{Automated} & \textbf{Query (gist)} & \textbf{Report A (chosen) vs B (rejected)} & \textbf{Wrong-judge's key claim} & \textbf{Why this category} \\
\midrule

\textbf{RMI} \newline {\scriptsize id=120 (RB)}
& 63.5\%
& 61.5\%
& RL techniques for small LLMs vs distillation
& \textbf{A}: \texttt{gpt-4o-search-preview} 55.6; 21 refs cycling only DistilBERT $+$ 1 blog. \newline \textbf{B}: \texttt{Perplexity} 33.3; 32 unique refs incl.\ Mar.\ 2025 arXiv.
& \textit{All 3 judges:} ``B has 2025 arXiv\,/\,GitHub citations; A repeats DistilBERT excessively. B's quantifiable results (AMC23 63$\to$80\%, \$42 cost) support its claims.''
& \textbf{Judge's axis:} B's RL freshness + concrete RL benchmarks (2025 arXiv, AMC23 63$\to$80\%, \$42 cost) --- only the benchmark maps to a rubric point (\emph{direct RL works on 7B}, w=2/18). \textbf{Rubric's axis:} ``\emph{why distillation beats RL}'' (3 sub-points: superiority-rationale, pre-optimized transfer, complexity-circumvention; \textbf{w=6/18}) --- all covered by A, all missed by B. Judge and rubric are scoring different things. \\
\midrule

\textbf{EKE} \newline {\scriptsize id=40 (DRB)}
& 7.7\%
& 12.1\%
& Top-10 Chinese box-office films $+$ future high-grossing types
& \textbf{A}: \texttt{doubao-deepresearch} 46.4. \newline \textbf{B}: \texttt{kimi-researcher} 38.9. \newline \emph{Both reports cite ``154.46\,yi yuan'' verbatim.}
& \textit{Qwen3:} ``154.46 billion RMB $\approx$ \$21.5B, over 5$\times$ Avatar's global box office. Clearly fabricated.'' \textit{Gemini:} ``Astronomically high\ldots\ far exceed any real-world record globally.''
& Judge doesn't know about \textbf{Nezha 2} (Jan.\ 2025 real record, \$2.1B), and misreads the Chinese unit \emph{yi} ($10^{8}$) as \emph{billion} ($10^{9}$), inflating the figure to \$21.5B. A real record becomes ``fabricated'' from a judge knowledge gap --- \textbf{Report B cites the same number but is not flagged}. \\
\midrule

\textbf{CEB} \newline {\scriptsize id=139 (RB)}
& 7.7\%
& 8.8\%
& Architectural innovations for lifelong learning in neural networks
& \textbf{A}: \texttt{Perplexity} 57.1; 40 real references. \newline \textbf{B}: \texttt{Claude} 50.0; 0 references.
& \textit{Gemini:} ``B highlights breakthroughs with specificity, e.g., `CH-HNN in Nature Communications 2025' with `60.82\% power reduction' \ldots \textbf{even if the specific citations are not real}.''
& Report B has \textbf{0 citations} yet produces specific-looking artifacts (``60.82\% power reduction'', ``Nature Communications 2025''). Judge treats numerical precision as evidence, \emph{explicitly} flagging the citations as possibly fictional yet still picking B. \\
\midrule

\textbf{SLB} \newline {\scriptsize id=19 (DRB)}
& 7.7\%
& 5.5\%
& Anthropic's Streamable HTTP engineering implementation
& \textbf{A}: \texttt{grok-deeper-search} 51.4; 6.2\,KB; cites the canonical MCP spec. \newline \textbf{B}: \texttt{kimi-researcher} 43.8; 27.0\,KB (4$\times$ longer); benchmark from a single CNode blog.
& \textit{DeepSeek:} ``B provides architecture diagrams, protocol layer breakdowns, performance benchmarks (175\% throughput, 74\% memory reduction), concrete code snippets, clearer section headings.''
& Every deciding factor is \emph{structural} --- diagrams, code blocks, headings, B is 4$\times$ longer. The lone numeric benchmark traces to a non-authoritative blog; A cites the \textbf{canonical MCP spec}. Judge equates surface organization with depth. \\
\midrule

\textit{Other} & 13.5\% & 12.1\% & \multicolumn{4}{@{}>{\raggedright\arraybackslash}p{11.8cm}@{}}{\textit{ (empty CoT, 7/7 from Gemini-2.5-Flash) . No representative case shown.}} \\

\bottomrule
\end{tabular}

\caption{Error taxonomy with representative cases. \textbf{Manual}: human annotation on 52 wrong judgments drawn from 20 hand-selected hard cases. \textbf{Automated}: few-shot LLM classification applied to every wrong judgment from the generative judges. Judge concentration is preserved across both views: EKE clusters on Qwen3-235B, SLB on DeepSeek-V3, and the empty-CoT pathology exclusively on Gemini-2.5-Flash; RMI and CEB are spread across all three judges. CoT excerpts are verbatim. RB = Researcher Bench; DRB = Deep Research Bench. Scores are RACE recall (out of 100).}
\label{tab:error-taxonomy-cases}
\end{table*}
\clearpage
\twocolumn

\end{document}